\documentclass[11pt]{article}

\usepackage[final]{acl}

\usepackage{times}
\usepackage{latexsym}
\usepackage[T1]{fontenc}
\usepackage[utf8]{inputenc}
\usepackage{microtype}
\usepackage{inconsolata}
\usepackage{graphicx}

\usepackage{hyperref}
\usepackage{url}
\usepackage{enumitem}
\usepackage{amsmath}
\usepackage{adjustbox}
\usepackage{booktabs}
\usepackage{multirow}
\usepackage{tabularx}
\usepackage{array}
\usepackage{makecell}
\usepackage{float}
\usepackage{caption}
\usepackage{subcaption}
\usepackage{stfloats}

\usepackage[table, svgnames]{xcolor}
\definecolor{darkgray}{RGB}{90,90,90}
\colorlet{red}{red!50!white}
\colorlet{yellow}{yellow!50!white}
\colorlet{blue}{SkyBlue!50!white}
\colorlet{lightred}{red!90!white}
\colorlet{lightyellow}{yellow!90!white}
\colorlet{lightblue}{SkyBlue!90!white}
\colorlet{lightorange}{orange!80!white}
\colorlet{lightgreen}{LimeGreen!50!white}
\colorlet{lightpurple}{purple!40!blue!60!white}

\usepackage{tcolorbox}
\usepackage{arydshln}

\newcolumntype{S}{>{\raggedright\arraybackslash}m{1.46cm}}
\newcolumntype{H}{>{\raggedright\arraybackslash}m{4.09cm}}
\newcolumntype{C}{>{\centering\arraybackslash}m{0.79cm}}
\newcolumntype{M}{>{\centering\arraybackslash}m{0.68cm}}
\newcolumntype{G}{!{\color{gray!50}\vrule width 0.5pt}}
\newcommand{\grayhline}{\arrayrulecolor{gray!30}\hline\arrayrulecolor{black}}

\providecommand{\tabgain}[2]{%
#1%
\rlap{\hspace{-1.75em}\raisebox{-0.8em}{%
\tiny\textcolor{green!70!black}{#2}}}
}
\providecommand{\tabloss}[2]{%
#1%
\rlap{\hspace{-1.75em}\raisebox{-0.8em}{%
\tiny\textcolor{darkgray}{#2}}}
}

\newcommand{\tabgainsubscript}[2]{%
  #1%
  \textsubscript{\textcolor{green!70!black}{\tiny #2}}%
}
\providecommand{\tablosssubscript}[2]{
  #1%
  \textsubscript{\textcolor{darkgray}{\tiny #2}}%
}
\providecommand{\tabsamesubscript}[2]{
  #1%
  \textsubscript{\textcolor{darkgray}{\tiny #2}}%
}

\renewcommand{\footnotesize}{\scriptsize}

\usepackage{silence}
\title{T2T-VICL: Cross-Task Visual In-Context Learning \\via Implicit Text-Driven VLMs}

\author{
Shao-Jun Xia\thanks{Equal contribution.}\\
Duke University\\
{\texttt\small shaojun.xia@duke.edu}
\\\And
Huixin Zhang\footnotemark[1]\\
Texas A\&M University\\
{\texttt\small zhanghui21@tamu.edu}
\\\And
Zhengzhong Tu\thanks{Corresponding author.}\\
Texas A\&M University\\
{\texttt\small tzz@tamu.edu}
}

\begin{document}
\maketitle
\begin{abstract}
Visual in-context learning (VICL) solves visual tasks by conditioning on a few input-output demonstrations without any model training.
Recent advances in large vision-language models (VLMs) have shown promising VICL capability when the demonstration pair and the query belong to the same vision task, but real use cases often provide mismatched examples, making it unclear whether a VLM should imitate the demonstrated transformation or infer a new one from the query.
This raises a fundamental question:
\textbf{\textit{Can VLMs perform cross-task VICL where demonstration and query differ?}} 
In this paper, we study this cross-task VICL setting and propose T2T-VICL, a collaborative prompt-transfer framework that converts mismatched visual demonstrations into implicit textual guidance without explicitly naming the tasks.
To do so, a large teacher VLM first generates structured descriptions of visual changes and task differences between task pairs, from which we construct a dataset of diverse implicit cross-task relations. We then distill this capability into a lightweight student VLM that produces content-dependent prompts from a task-A demonstration pair and a task-B query. The generated prompt is used to guide a frozen image-editing VLM.
Experiments on 12 low-level vision tasks and over 20 evaluated cross-task pairs show that T2T-VICL consistently improves task-aware alignment over fixed prompting and often also improves image fidelity, revealing both the potential and limits of cross-task VICL. Our code is available on GitHub.\footnote{\url{https://github.com/ZhangHuixin1103/Task-Transfer-VICL-VLMs}}
\end{abstract}

\section{Introduction}
\label{sec:intro}

In-context learning (ICL) allows foundation models, especially large language models (LLMs), to adapt to new tasks from only a few input-output demonstrations, without updating model parameters~\cite{min2022rethinking, dong2024survey, highmore2024context, wang2023large, sia2024does, li2025large}.
Recent progress in large vision-language models (VLMs) has extended this paradigm from language to multimodal settings, giving rise to visual in-context learning (VICL), where a model infers an image transformation directly from a small number of visual examples~\cite{bar2022visual, wang2023images, wang2023context, sun2024x, zhou2024visual, ma2025vision}.
Such a capability is particularly appealing for low-level vision, where one would ideally like a single frozen model to handle diverse image restoration, removal, and generation tasks by simply conditioning on demonstrations rather than retraining for each task.

However, existing VICL methods largely rely on a restrictive \emph{same-task} assumption: the demonstration pair and the query are usually drawn from the same task. In practice, this assumption is often violated. For example, a user may only have a \emph{deraining} example while the query requires \emph{dehazing}, or may provide a reflection-removal pair while the true target is shadow removal. This mismatch makes VICL fundamentally ambiguous: should the model imitate the demonstrated transformation, or infer a different one from the query itself? At the same time, many low-level tasks are not isolated~\cite{zamir2018taskonomy, pal2019zero, achille2019task2vec, bao2019information}. They share latent relationships in terms of visual effects, such as artifact suppression, contrast recovery, color adjustment, or illumination correction. Importantly, these relationships are often easier to express in language than in raw pixels, suggesting that language may serve as a natural bridge for transferring knowledge across mismatched visual tasks~\cite{trigka2025comprehensive, shu2024visual, brooks2023instructpix2pix, potlapalli2023promptir, conde2024instructir, ke2023neural}.

\begin{figure*}[ht]
    \centering
    \includegraphics[width=0.89\textwidth]{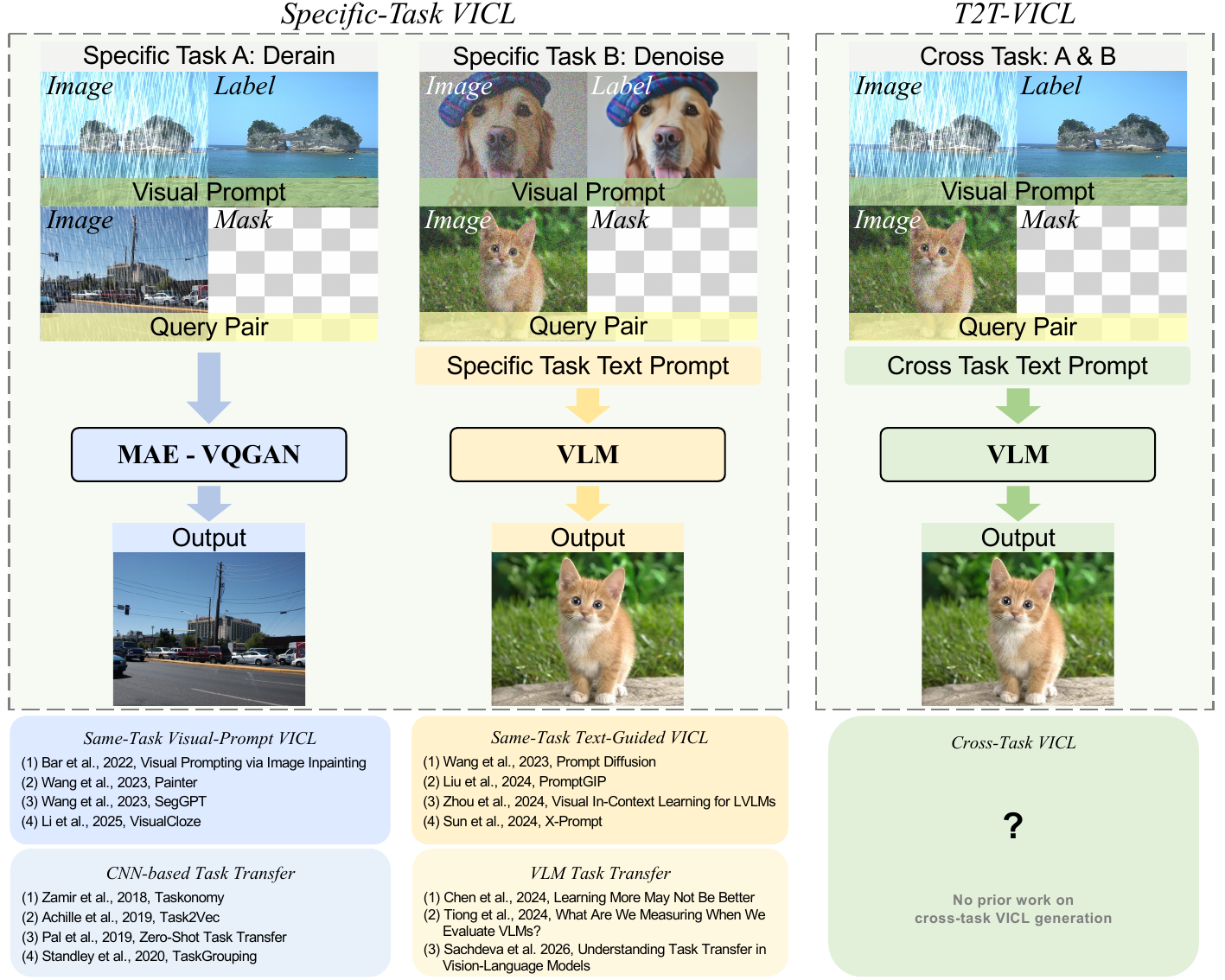}
    \caption{Illustration of Cross-Task Visual In-Context Learning}
    \label{fig:fig1}
\vspace{-11pt}
\end{figure*}

These observations motivate a new question: \emph{Can VLMs perform visual in-context learning when the demonstration and the query come from different tasks?} We study this setting as \textbf{cross-task visual in-context learning}. Compared with standard VICL, cross-task VICL requires more than pattern imitation. The model must abstract the visual change encoded in the context, reason about the discrepancy between the demonstrated task and the query task, and then produce the correct target transformation without being explicitly told the task name. A fixed prompt is often insufficient for this setting, because it provides only a generic instruction while leaving the task difference unresolved. What is needed is a mechanism that converts mismatched visual demonstrations into task-aware guidance that remains compatible with frozen VLM generators.
Figure~\ref{fig:fig1} contrasts conventional same-task VICL and traditional task transfer with the cross-task VICL setting studied here, illustrating the gap addressed by T2T-VICL.

To this end, we propose \textbf{T2T-VICL}, a collaborative teacher-student prompting framework for cross-task VICL. Our key idea is to translate the relationship between two tasks into \emph{implicit task-difference prompts} that describe visual changes rather than explicitly name tasks. A large teacher VLM first generates structured implicit descriptions from cross-task visual prompts, and we curate these outputs into a benchmark of cross-task text-image relations. This capability is then distilled into a lightweight student VLM, which produces content-dependent prompts from a task-$A$ demonstration pair and a task-$B$ query. The generated prompt is fed to a frozen image-editing VLM, while an automatic score-based inference strategy, combining VLM-based visual instruction scoring~\cite{ku2024viescore} and image quality assessment (IQA) metrics~\cite{zhang2012comprehensive}, is used to assess generated outputs. In this way, T2T-VICL provides a practical framework for probing and improving cross-task VICL without updating the final generator model.
We then evaluate T2T-VICL across diverse cross-task pairs and image-editing VLMs using quantitative metrics and human preference studies, and further compare other visual prompting and image-editing methods with ours. The results show that T2T-VICL can improve task-aware alignment and achieve competitive image quality across cross-task pairs.

Our contributions are three-fold:
\begin{itemize}[itemsep=-2pt, topsep=1pt]
    \item  We formulate \textbf{cross-task VICL} for low-level vision and introduce a benchmark with implicit cross-task descriptions, enabling systematic study of how VLMs generalize across mismatched visual demonstrations. 
    \item  We propose T2T-VICL, a \textbf{VLM $\rightarrow$ sVLM $\rightarrow$ VLM} teacher-student pipeline that distills implicit task-relation knowledge from a large model into a lightweight prompt generator for frozen image-editing VLMs. 
    \item  We develop a \textbf{score-based inference framework} that combines visual instruction scoring and IQA metrics, providing an automatic way to support and evaluate cross-task VICL.
\end{itemize}

\section{Related Work}
\label{sec:relat}

\subsection{Vision Foundation Models}

\noindent Vision foundation models aim to unify diverse visual tasks within a single framework. Early approaches reformulated vision problems as sequence prediction or instruction-based learning (Pix2Seq~\cite{chen2022unified}, Unified-IO~\cite{lu2022unified}, OFA~\cite{wang2022ofa}), while later works introduced task-agnostic multimodal interfaces (UViM~\cite{kolesnikov2022uvim}, Uni-Perceiver~\cite{zhu2022uni}). Recent advances further leverage stronger multimodal backbones and instruction-driven paradigms for prompt-based recognition and conditional generation (Florence-2~\cite{xiao2024florence}, InstructCV~\cite{gan2024instructcv}, InstructDiffusion~\cite{geng2024instructdiffusion}). Meanwhile, VLMs have emerged as powerful vision generalists through unified multimodal architectures, including Emu~\cite{sun2024emu,wang2024emu3}, Chameleon~\cite{team2024chameleon}, Transfusion~\cite{zhou2025transfusion}, and Show-o~\cite{xie2025show}. Collectively, these efforts highlight the trend of VLMs shaping the evolution of vision generalist models.

\subsection{Visual In-Context Learning}
Visual in-context learning (VICL) refers to adapting vision models to downstream tasks through contextual examples rather than explicit fine-tuning, and it can be broadly divided into visual prompting and prompt-driven conditioning. Implicit prompting methods, represented by MAE-VQGAN~\cite{bar2022visual} and Painter~\cite{wang2023images}, rely on masked prediction or image completion as objectives, enabling models to generalize across diverse vision tasks. In contrast, explicit prompt-driven approaches leverage explicit conditioning to adapt to diverse vision problems. For instance, Prompt Diffusion~\cite{wang2023context} exploits diffusion-based generation under prompt guidance, while PromptGIP~\cite{liu2024unifying} adopts QA-style prompt structures. ImageBrush~\cite{yang2023imagebrush} learns exemplar-based visual instructions for image manipulation, and InstructDiffusion~\cite{geng2024instructdiffusion} instead consumes a query image and language instruction. Additional works such as CoOp~\cite{zhou2022learning} and VPT~\cite{jia2022visual} further extend VICL by introducing learnable prompts for vision transformers. More recently, X-Prompt~\cite{sun2024x} targets general image generation, compressing contextual signals and unifying diverse vision tasks within an auto-regressive framework, and VisualCloze~\cite{li2025visualcloze} further studies universal image generation via visual demonstrations, where task instructions are inferred from example input-output pairs rather than provided only through language.
Most VICL methods assume that the visual prompt and query image belong to the same task, while our T2T-VICL extends this paradigm to difficult cross-task settings, where the visual prompt and query image are from different tasks, unlocking the potential for knowledge adaptation across diverse vision tasks.

\subsection{Task Transfer}
Task generalization refers to leveraging knowledge from previously solved tasks to facilitate learning on unseen ones. Early milestones such as Taskonomy~\cite{zamir2018taskonomy} provided the first large-scale analysis of transferability across 26 vision tasks, introducing the task affinity matrix as a foundation for cross-task generalization. Following this line, TTNet~\cite{pal2019zero} leveraged inter-task correlations to predict model parameters for unseen tasks without requiring target-task labels. Similarly, Task2Vec~\cite{achille2019task2vec} proposed embedding tasks into a vector space using Fisher information, making task similarity measurable and interpretable. Beyond these foundations, \citet{standley2020tasks} systematically investigated which tasks should be learned together in multitask settings, highlighting task affinity as a guiding principle. RSA~\cite{dwivedi2019representation} introduced representation similarity to estimate inter-task relationships. Recent work has extended transfer analysis to VLMs. \citet{chen2024learning} conducted a study of knowledge transferability across vision-language tasks, showing that sequential transfer can be either beneficial or detrimental depending on task grouping, dataset scale, and the pre-training stage. \citet{tiong2024we} analyzed transfer behavior across training and evaluation tasks, using the transfer matrix to uncover latent vision-language capabilities and reveal biases in existing evaluation benchmarks. More recently, \citet{sachdeva2026understanding} studied task transfer in VLMs by measuring how fine-tuning on one perception task affects zero-shot performance on others, constructing task-transfer graphs from the resulting transfer behavior.
The studies above characterize transfer induced by task-specific training, whereas our work asks whether an in-context visual demonstration from one task can support image generation for a different task at inference time.
Given the growing visual generation ability of VLMs, the progress in this field motivates our exploration of cross-task generalization in VLMs and unlocks the boundaries.

\subsection{Language-Driven Image Editing}
\noindent Language-driven vision restoration has evolved from instruction-guided editing to adaptive prompt-based restoration. Early works such as InstructPix2Pix~\cite{brooks2023instructpix2pix} and InstructIR~\cite{conde2024instructir} showed that natural language can guide fine-grained, degradation-aware manipulation of image features. Subsequent studies including PromptIR~\cite{potlapalli2023promptir}, SPIRE~\cite{qi2024spire}, PromptFix~\cite{yu2024promptfix}, Perceive-IR~\cite{zhang2025perceive}, and FPro~\cite{zhou2024seeing} further enriched semantic and perceptual prompting, enabling finer control over fidelity and restoration strength. These advances establish natural language as a direct and interpretable driver for low-level vision tasks.

\section{Framework}
\label{sec:method}

\begin{figure*}[t]
    \centering
    \includegraphics[width=0.9\textwidth]{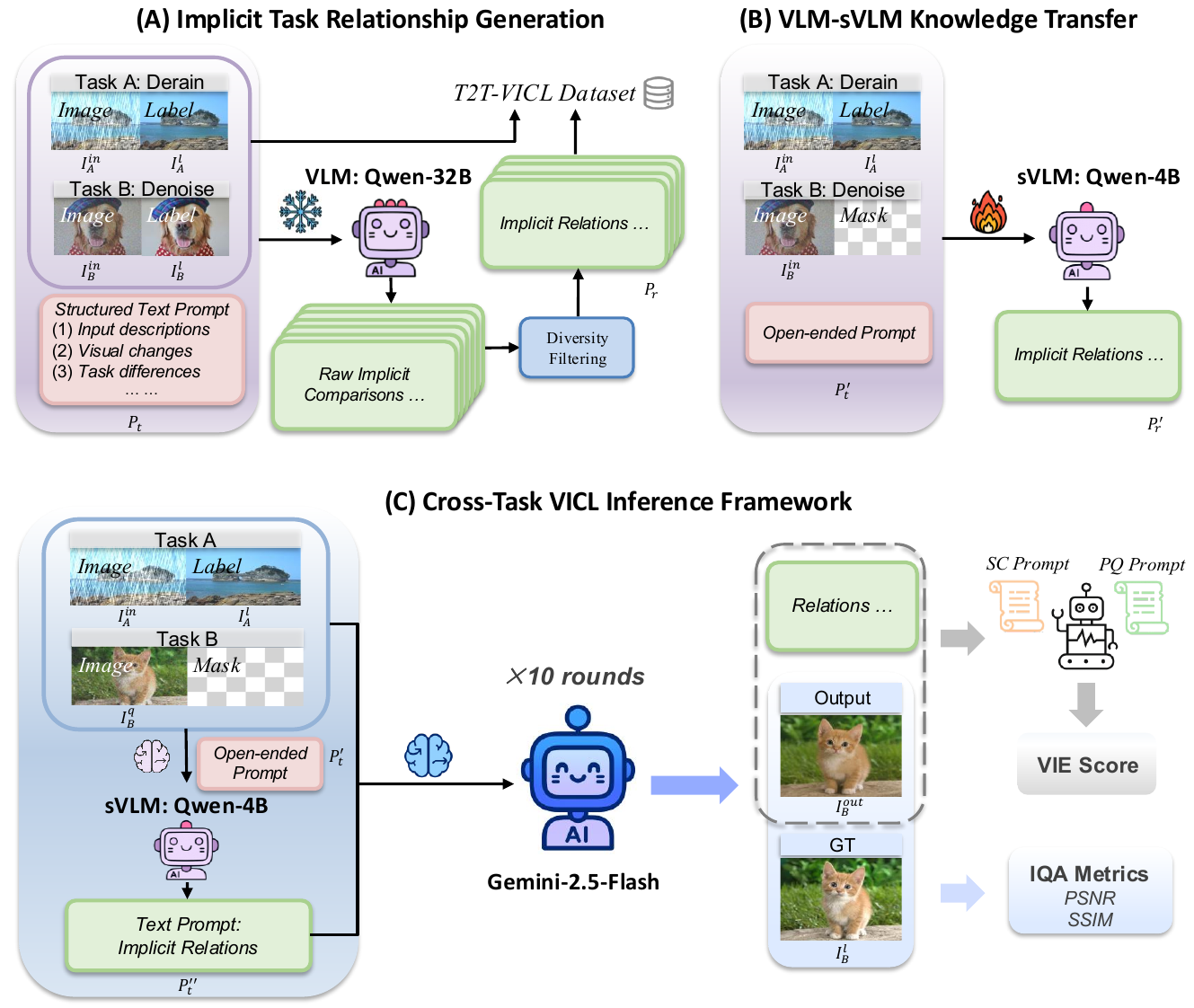}
    \caption{Overview of the Proposed Pipeline and Its Workflow}
    \label{fig:fig2}
\vspace{-8pt}
\end{figure*}

\subsection{T2T-VICL: Overview of VLM Pipeline}
To explore the boundaries of cross-task VICL via VLMs, we propose T2T-VICL, a collaborative pipeline that leverages the complementary strengths of large and small VLMs to enhance vision-language reasoning. First, a large pre-trained VLM generates implicit textual descriptions for two distinct tasks, while the linguistic reliability of these descriptions is quantitatively evaluated. To enable efficient deployment, we fine-tune an sVLM to transfer the reasoning capabilities of the large model while reducing computational overhead. During inference, the sVLM produces text prompts for the final frozen VLM, thereby guiding the downstream reasoning process. Our main VLM inference pipeline aligns with the standard VICL setting, while the trained sVLM is only an auxiliary component that extracts implicit task descriptions from visual examples and generates text prompts without altering the main inference branch. Finally, generated results are scored by another VLM with a visual instruction-guided metric and IQA metrics.

\subsection{Implicit Relationship Generation}
\label{sec:teacher}

We begin by automatically generating rich textual descriptions that implicitly capture the differences between two low-level vision tasks. We divide 12 diverse low-level tasks into three categories, as shown in Table~\ref{tab:1}, spanning classic degradation problems (deraining, dehazing, denoising, deblurring, demoireing), removal tasks (shadow removal, reflection removal) as well as generation/enhancement tasks (colorization, low-light enhancement, harmonization, inpainting, style transfer). For any arbitrary pair of tasks $A$ and $B$, our goal is to obtain an implicit text prompt that depicts the difference without explicitly naming either task.
To accomplish this, we leverage the large vision-language model Qwen2.5-VL-32B-Instruct, denoted by $M_t$, which takes two image pairs as input. We design the text prompt $P_t$ to provide structured comparisons between two tasks by incorporating (1) image descriptions representing the basic content of the images from tasks $A$ and $B$, (2) visual changes highlighting observable image-level changes rather than task labels, and (3) task differences specifying the perceptible modifications made by each task.
Crucially, the prompt instructs the VLM to compare these aspects without explicitly stating either task name or telling the model what exactly the tasks should do, ensuring that the description remains extremely implicit. This mechanism forces the VLM to articulate the subtle differences in a narrative manner.
Meanwhile, we also consider the baseline setting, where the VLM performs VICL with a fixed prompt. Details of the implicit and fixed prompts are provided in Appendix Sec.~\ref{sec:prompt}.



\begin{table}[!ht]
\vspace{-6pt}
\centering
\footnotesize
\captionsetup{width=0.99\columnwidth, skip=4pt, type=table}
\caption{Categorization of low-level vision tasks. Based on this, we divide cross-task pairs into \textbf{intra-category} compositions such as ``Deblurring $\rightarrow$ Demoireing'', and \textbf{inter-category} compositions such as ``Reflection Removal $\rightarrow$ Dehazing''.}
\vspace{-1pt}
\renewcommand{\arraystretch}{1}
\begin{tabularx}{\columnwidth}{p{0.24\columnwidth} X}
\hline
\textbf{Category} & \textbf{Tasks} \\ \hline
\rowcolor{lightred!30}
\textbf{Restoration} & \makecell[l]{Deblurring, Dehazing, Demoireing, \\Denoising, Deraining} \\
\rowcolor{lightyellow!30}
\textbf{Removal} & Reflection Removal, Shadow Removal \\
\rowcolor{lightblue!30}
\textbf{Generation\slash Enhancement} & Colorization, Harmonization, Inpainting, Light Enhancement, Style Transfer \\
\hline
\end{tabularx}
\label{tab:1}
\vspace{-8pt}
\end{table}

Subsequently, the above procedure is implemented for all the included low-level tasks to build a high-quality benchmark dataset. For each pair of tasks $A$ and $B$, we sample a large number of combinations from public datasets and query the VLM with each sample to get a text output: (i) an input image $I_A^{in}$ and its ground-truth label image $I_A^{l}$ from task $A$, and (ii) an input image $I_B^{in}$ with the label $I_B^{l}$ from a different task $B$. We define this ordered image tuple as $\mathcal{X}_t$ and obtain the implicit textual relation $P_r$:
\begin{equation}
\begin{aligned}
\mathcal{X}_t &= (I_A^{in}, I_A^{l}, I_B^{in}, I_B^{l}),\\
P_r &= M_t(\mathcal{X}_t, P_t).
\end{aligned}
\label{eq:teacher-input}
\end{equation}
The teacher model generates the corresponding comparison text for each pairing. However, the VLM can sometimes produce repetitive or overly generic statements, especially if many samples share similar traits. We therefore conduct a diversity filtering method by using semantic sentence embeddings. We encode each candidate description with a SentenceTransformer~\cite{reimers2019sentence} (all-MiniLM-L6-v2) to obtain a dense vector representation. This allows us to quantitatively measure similarities among the descriptions. We then perform clustering in embedding space to detect groups of near-duplicates or redundant phrasing. In this manner, we filter out repetitive outputs and retain one representative description from each cluster. Ultimately, we keep 2,000 diverse descriptions per task pair. To our knowledge, this is the first text-and-image dataset that implicitly captures cross-task relationships, providing the foundation for our next steps.

\subsection{VLM-sVLM Knowledge Transfer}

\subsubsection{Large-to-Small Transfer Framework}

Having obtained the implicit task comparisons from the large VLM, we next transfer this knowledge into a Qwen3-VL-4B-Instruct model as the student $M_s$. Using the generated text prompts in Sec.~\ref{sec:teacher} as targets, we fine-tune the student with an ordered image tuple $\mathcal{X}_s$, removing the label image $I_B^{l}$ from $\mathcal{X}_t$. Given an open-ended prompt $P_t'$ (e.g., \textit{Compare the observations in these images}), we force the student model to learn to infer the effect of task $B$ from the characteristics of input and output the implicit textual relation $P_r'$.
\begin{equation}
\begin{aligned}
\mathcal{X}_s &= (I_A^{in}, I_A^{l}, I_B^{in}),\\
P_r' &= M_s(\mathcal{X}_s, P_t').
\end{aligned}
\label{eq:student-input}
\end{equation}
We format each training instance in the official Qwen-VL conversational style, with image placeholders in the user message. The student is then trained to produce the teacher's full comparison text as the completion. The training objective is to minimize the cross-entropy loss between the generated text from the student and the reference description from the teacher. Through many epochs of exposure to different task pairs and images, the sVLM gradually learns to reproduce the teacher's task-comparison behavior and becomes capable of generating such comparisons on its own. In effect, the implicit knowledge of inter-task relationships is transferred into the small model at the language level.

We selected Qwen3-VL-4B as the student not only for its manageable size, but also because it has an architecture and multimodal interfaces similar to those of the teacher model. After training, the fine-tuned 4B model is indeed able to take a task-$A$ image pair and a task-$B$ query and directly output a coherent comparison of the tasks, much like the 32B model. In other words, the student has learned to generate content-dependent prompts that describe cross-task relationships on demand. This approach is related in spirit to recent prompt-based tuning methods in vision-language models~\cite{zhou2022learning}. But the key difference is that instead of learning a static prompt vector or fixed set of words for a model, we train a full generative model to produce dynamic, content-dependent prompts. The student can flexibly generate different comparative descriptions for different image pairs, rather than a single fixed prompt. By using the outputs of the large model as textual supervision, we effectively compress its high-level reasoning about tasks into a lightweight model.

\subsubsection{Small-to-Large Deployment Framework}

Once the student model is trained, we deploy it as a front-end module that provides a text prompt to the larger VLM for final reasoning and output image generation. This hierarchical approach reduces reliance on heavy computation during the initial stages of inference while preserving the representational capacity of large-scale VLMs. This design inverts the conventional direction of knowledge flow, where the Small-to-Large step leverages the student model to guide the final inference model (Gemini-2.5-Flash). The process employs the trained Qwen3-VL-4B model to process the input of the visual prompt ($I_A^{in}$, $I_A^{l}$), query image $I_B^{q}$, and text prompt $P_t'$, and output a prompt representation $P_t''$ as follows:
\begin{equation}
P_t'' = M_s(I_A^{in}, I_A^l, I_B^q, P_t').
\end{equation}
By offloading this initial abstraction to $M_s$, we reduce the computational overhead associated with feeding raw multimodal inputs directly into the large model.
$P_t''$ has the same form as the implicit relation $P_r'$ produced during training.

The generated prompt $P_t''$ is then provided as an input to a larger VLM, denoted by $M_\ell$, which possesses stronger representational power and reasoning ability. Specifically,
\begin{equation}
I_B^{out} = M_\ell(I_A^{in}, I_A^l, I_B^q, P_t''),
\end{equation}
where $I_B^{out}$ denotes the model prediction. This step enables $M_\ell$ to focus on higher-level reasoning and executing the described transformation on the query image, instead of figuring out what the transformation should be. More importantly, this two-stage deployment is highly interpretable, as the intermediate text prompt $P_t''$ clearly explains the intended operation. It also offers a point of control at which a human or another module could modify or validate the prompt before final image generation. This pipeline leverages the complementary strengths of each model size, combining the powerful reasoning and generation of the large VLM with the efficiency and fast text generation of the sVLM to achieve effective cross-task visual in-context learning.


\subsubsection{Score-Based Reasoning}

To enhance decision-making, we introduce perceptual score-based evaluation based on VIEScore~\cite{ku2024viescore}. Since conventional image-synthesis metrics are often task-agnostic and opaque, they provide a single number without revealing why an image is judged good or bad. In contrast, VIEScore is a task-aware and explainable evaluator driven by a VLM. Given an instruction $I$, a synthesized image $O$, and a set of conditions $C^*$, the evaluator first produces a natural-language rationale and then a scalar score $s$:
\begin{equation}
f_{\text{VIE}}(I,O,C^{*}) = (\text{rationale},\, s).
\end{equation}
Since a generated image should be penalized when either instruction following or visual quality fails, we separately evaluate \emph{Semantic Consistency} (SC) and \emph{Perceptual Quality} (PQ).
Each is formed by minimum aggregation over task-specific sub-scores $\{\alpha_i\}$ and $\{\beta_j\}$ on a 0--10 scale:
\[
\mathrm{SC}=\min_i \alpha_i,\qquad \mathrm{PQ}=\min_j \beta_j.
\]
The final overall rating uses a geometric mean:
\begin{equation}
s=\big(\mathrm{SC}\cdot \mathrm{PQ}\big)^{1/2}.
\end{equation}
In practice, we prompt the VLM with explicit rubrics for SC and PQ. Notably, PQ is assessed from the synthesized image alone (to avoid instruction confounds), while SC conditions are presented alongside the image. 
This design yields interpretable rationales, task-aware scores, and robust correlations with human judgments across diverse conditional image tasks.

\begin{table*}[ht]
\centering
\scriptsize
\setlength{\tabcolsep}{0.5pt}
\renewcommand{\arraystretch}{0.96}
\captionsetup{width=0.98\textwidth, skip=4pt, type=table}
\caption{Average of ground-truth-based best-of-10 evaluation results in the top-tier group. Generated by Gemini and Seedream with Qwen prompts, compared to the fixed-prompt baseline.}
\vspace{-1pt}
\label{tab:3}

\begin{adjustbox}{width=\textwidth}
\begin{tabular}{>{\raggedright\arraybackslash\leftskip=3pt}m{\dimexpr4.09cm+3pt\relax} *{12}{>{\centering\arraybackslash}m{0.84cm}}}
\toprule
\multirow{3}{*}{\textbf{Task}}
& \multicolumn{6}{c}{\textbf{Gemini 2.5 Flash}}
& \multicolumn{6}{c}{\textbf{Seedream 4.0}} \\
& \multicolumn{2}{c}{\textbf{PSNR}$\uparrow$}
& \multicolumn{2}{c}{\textbf{SSIM}$\uparrow$}
& \multicolumn{2}{c}{\textbf{VIEScore}$\uparrow$}
& \multicolumn{2}{c}{\textbf{PSNR}$\uparrow$}
& \multicolumn{2}{c}{\textbf{SSIM}$\uparrow$}
& \multicolumn{2}{c}{\textbf{VIEScore}$\uparrow$} \\
& \textcolor{gray}{\textbf{Fixed}} & \textcolor{gray}{\textbf{Ours}}
& \textcolor{gray}{\textbf{Fixed}} & \textcolor{gray}{\textbf{Ours}}
& \textcolor{gray}{\textbf{Fixed}} & \textcolor{gray}{\textbf{Ours}}
& \textcolor{gray}{\textbf{Fixed}} & \textcolor{gray}{\textbf{Ours}}
& \textcolor{gray}{\textbf{Fixed}} & \textcolor{gray}{\textbf{Ours}}
& \textcolor{gray}{\textbf{Fixed}} & \textcolor{gray}{\textbf{Ours}} \\
\midrule

\rowcolor{lightred!30}
Deblurring $\rightarrow$ Dehazing
& 11.47 & \cellcolor{lightred!60}\tabgain{12.35}{+0.88}
& 0.473 & \cellcolor{lightred!60}\tabgain{0.511}{+0.038}
& 6.40 & \cellcolor{lightred!60}\tabgain{7.61}{+1.21}
& 10.50 & \cellcolor{lightred!60}\tabgain{11.08}{+0.58}
& 0.345 & \cellcolor{lightred!60}\tabgain{0.372}{+0.027}
& 8.22 & \cellcolor{lightred!60}\tabgain{8.48}{+0.26} \\
\grayhline

\rowcolor{lightred!30}
Deblurring $\rightarrow$ Deraining
& 19.21 & \cellcolor{lightred!60}\tabgain{19.31}{+0.10}
& 0.516 & \cellcolor{lightred!60}\tabgain{0.527}{+0.011}
& 7.04 & \cellcolor{lightred!60}\tabgain{7.85}{+0.81}
& \cellcolor{lightred!60}14.68 & \tabloss{13.62}{-1.06}
& 0.371 & \cellcolor{lightred!60}\tabgain{0.377}{+0.006}
& 6.27 & \cellcolor{lightred!60}\tabgain{7.35}{+1.08} \\
\grayhline

\rowcolor{lightred!30}
Deblurring $\rightarrow$ Demoireing
& 14.45 & \cellcolor{lightred!60}\tabgain{14.60}{+0.15}
& \cellcolor{lightred!60}0.606 & \tabloss{0.605}{-0.001}
& 5.70 & \cellcolor{lightred!60}\tabgain{6.25}{+0.55}
& 10.99 & \cellcolor{lightred!60}\tabgain{11.45}{+0.46}
& 0.435 & \cellcolor{lightred!60}\tabgain{0.477}{+0.042}
& 6.56 & \cellcolor{lightred!60}\tabgain{7.36}{+0.80} \\
\grayhline

\rowcolor{lightred!30}
Demoireing $\rightarrow$ Dehazing
& 11.20 & \cellcolor{lightred!60}\tabgain{12.40}{+1.20}
& 0.478 & \cellcolor{lightred!60}\tabgain{0.533}{+0.055}
& 5.31 & \cellcolor{lightred!60}\tabgain{7.55}{+2.24}
& 11.15 & \cellcolor{lightred!60}\tabgain{11.24}{+0.09}
& \cellcolor{lightred!60}0.397 & \tabloss{0.392}{-0.005}
& 8.27 & \cellcolor{lightred!60}\tabgain{8.58}{+0.31} \\
\grayhline

\rowcolor{lightblue!30}
Harmonization $\rightarrow$ Light Enhancement
& 10.89 & \cellcolor{lightblue!60}\tabgain{12.81}{+1.92}
& 0.379 & \cellcolor{lightblue!60}\tabgain{0.518}{+0.139}
& 8.15 & \cellcolor{lightblue!60}\tabgain{8.65}{+0.50}
& 9.85 & \cellcolor{lightblue!60}\tabgain{10.05}{+0.20}
& 0.353 & \cellcolor{lightblue!60}\tabgain{0.365}{+0.012}
& 8.60 & \cellcolor{lightblue!60}\tabgain{8.67}{+0.07} \\
\grayhline

\rowcolor{lightblue!30}
Inpainting $\rightarrow$ Light Enhancement
& 10.88 & \cellcolor{lightblue!60}\tabgain{13.22}{+2.34}
& 0.377 & \cellcolor{lightblue!60}\tabgain{0.540}{+0.163}
& 7.35 & \cellcolor{lightblue!60}\tabgain{8.96}{+1.61}
& 9.35 & \cellcolor{lightblue!60}\tabgain{9.65}{+0.30}
& 0.340 & \cellcolor{lightblue!60}\tabgain{0.349}{+0.009}
& 8.50 & \cellcolor{lightblue!60}\tabgain{8.53}{+0.03} \\
\grayhline

\rowcolor{lightblue!30}
Inpainting $\rightarrow$ Style Transfer
& 9.70 & \cellcolor{lightblue!60}\tabgain{13.02}{+3.32}
& 0.340 & \cellcolor{lightblue!60}\tabgain{0.428}{+0.088}
& 1.77 & \cellcolor{lightblue!60}\tabgain{4.84}{+3.07}
& 9.75 & \cellcolor{lightblue!60}\tabgain{11.05}{+1.30}
& \cellcolor{lightblue!60}0.367 & \tabloss{0.310}{-0.057}
& 4.84 & \cellcolor{lightblue!60}\tabgain{6.60}{+1.76} \\
\grayhline

\rowcolor{lightblue!30}
Style Transfer $\rightarrow$ Light Enhancement
& 12.22 & \cellcolor{lightblue!60}\tabgain{14.33}{+2.11}
& 0.485 & \cellcolor{lightblue!60}\tabgain{0.571}{+0.086}
& 7.85 & \cellcolor{lightblue!60}\tabgain{8.74}{+0.89}
& 10.35 & \cellcolor{lightblue!60}\tabgain{10.61}{+0.26}
& 0.391 & \cellcolor{lightblue!60}\tabgain{0.406}{+0.015}
& 8.73 & \cellcolor{lightblue!60}\tabgain{8.79}{+0.06} \\
\grayhline

\rowcolor{lightpurple!30}
Denoising $\rightarrow$ Light Enhancement
& 10.23 & \cellcolor{lightpurple!60}\tabgain{12.42}{+2.19}
& 0.365 & \cellcolor{lightpurple!60}\tabgain{0.506}{+0.141}
& 7.27 & \cellcolor{lightpurple!60}\tabgain{8.18}{+0.91}
& 8.84 & \cellcolor{lightpurple!60}\tabgain{9.09}{+0.25}
& 0.314 & \cellcolor{lightpurple!60}\tabgain{0.358}{+0.044}
& 8.15 & \cellcolor{lightpurple!60}\tabgain{8.67}{+0.52} \\
\grayhline

\rowcolor{lightpurple!30}
Light Enhancement $\rightarrow$ Deraining
& 18.28 & \cellcolor{lightpurple!60}\tabgain{19.12}{+0.84}
& 0.472 & \cellcolor{lightpurple!60}\tabgain{0.522}{+0.050}
& 4.99 & \cellcolor{lightpurple!60}\tabgain{7.93}{+2.94}
& 11.89 & \cellcolor{lightpurple!60}\tabgain{13.16}{+1.27}
& 0.387 & \cellcolor{lightpurple!60}\tabgain{0.394}{+0.007}
& 6.46 & \cellcolor{lightpurple!60}\tabgain{7.86}{+1.40} \\
\grayhline

\rowcolor{lightgreen!30}
Light Enhancement $\rightarrow$ Shadow Removal
& 16.99 & \cellcolor{lightgreen!60}\tabgain{18.29}{+1.30}
& \cellcolor{lightgreen!60}0.344 & \tabloss{0.329}{-0.015}
& 5.99 & \cellcolor{lightgreen!60}\tabgain{8.04}{+2.05}
& 9.01 & \cellcolor{lightgreen!60}\tabgain{11.46}{+2.45}
& 0.309 & \cellcolor{lightgreen!60}\tabgain{0.329}{+0.020}
& 7.87 & \cellcolor{lightgreen!60}\tabgain{8.54}{+0.67} \\
\grayhline

\rowcolor{lightorange!30}
Reflection Removal $\rightarrow$ Dehazing
& 11.69 & \cellcolor{lightorange!50}\tabgain{12.37}{+0.68}
& 0.490 & \cellcolor{lightorange!50}\tabgain{0.514}{+0.024}
& 5.97 & \cellcolor{lightorange!50}\tabgain{7.82}{+1.85}
& \cellcolor{lightorange!50}11.16 & \tabloss{10.67}{-0.49}
& 0.358 & \cellcolor{lightorange!50}\tabgain{0.376}{+0.018}
& 8.27 & \cellcolor{lightorange!50}\tabgain{8.39}{+0.12} \\
\bottomrule
\end{tabular}
\end{adjustbox}
\vspace{-6pt}
\end{table*}

\subsubsection{Evaluation Protocols and Metrics}

To quantitatively assess our framework, we employ a hybrid metric suite. We adopt VIEScore for measuring the alignment between generated outputs and task-specific visual improvements, enabling evaluation from a reasoning perspective rather than pixel fidelity alone. For image quality, we report classical fidelity scores including Peak Signal-to-Noise Ratio (PSNR) and Structural Similarity Index (SSIM), which remain standard for restoration tasks. PSNR captures pixel-level fidelity via mean-squared error, whereas SSIM reflects perceptual alignment by jointly evaluating luminance, contrast, and structural consistency.
For some target tasks like colorization, style transfer, and deraining, where PSNR and SSIM may not fully capture perceptual quality, Appendix Sec.~\ref{sec:viescore} additionally reports CIEDE2000~\cite{luo2001development}, LPIPS~\cite{zhang2018unreasonable}, DISTS~\cite{ding2020image}, FID~\cite{heusel2017gans}, and ArtFID~\cite{wright2022artfid}.

For Gemini and Seedream, whose APIs can produce different outputs across calls, we use a \textit{best-of-10} protocol, as illustrated in Figure~\ref{fig:fig2}, and report a repeated-generation test in Appendix Sec.~\ref{sec:reliability}. We perform 10 independent generations for each query under both the fixed and the Qwen prompt, select the result with the highest PSNR as the final output, and then compute the corresponding SSIM and VIEScore.
Separately, for Gemini in Table~\ref{tab:2} and the additional open-source models in Table~\ref{tab:5}, we use a \textit{single-shot} protocol, performing one generation per query for both prompts and computing all metrics on that output without result selection.
Together, these metrics provide a holistic evaluation that balances low-level fidelity, perceptual similarity, and instruction alignment.


\section{Results}
\label{sec:result}

\subsection{Experimental Setup}
\label{sec:setup}

\textbf{Datasets.}
We build our study on a comprehensive collection of twelve representative low-level vision tasks. Specifically, we use GOPRO~\cite{nah2017deep} for image deblurring, D-HAZY~\cite{ancuti2016d} for dehazing, UHDM~\cite{yu2022towards} for demoireing, SIDD~\cite{abdelhamed2018high, abdelhamed2020ntire} for denoising, Rain13K~\cite{jiang2020multi,zamir2021multi} for deraining, iHarmony4~\cite{cong2019image} for harmonization, DIV2K~\cite{agustsson2017ntire} with synthetic masks for inpainting, modified ImageNet subset~\cite{deng2009imagenet,zhang2016colorful} for colorization, LoL~\cite{wei2018deep} for low-light enhancement, SIR$^2$~\cite{wan2022benchmarking} for reflection removal, ISTD~\cite{wang2018stacked} for shadow removal, and Night2Day~\cite{laffont2014transient} for style transfer. Details are provided in Appendix Sec.~\ref{sec:data}. The number of possible cross-task experiments scales quadratically with the number of tasks.

\noindent \textbf{Implementation details.} We conduct our experiments using two NVIDIA A100-SXM4 GPUs with 80 GB of memory each. We follow the standard train-test split in each dataset whenever available, or adopt a 70/30 random division when no official split is provided. All images are resized to a unified resolution ($448 \times 448$ for training inputs, $224 \times 224$ for query inputs) to maintain compatibility with the VLM framework.

\subsection{Cross-Task Results in Top-Tier Group}
\begin{figure*}[ht]
    \centering
    \includegraphics[width=\textwidth]{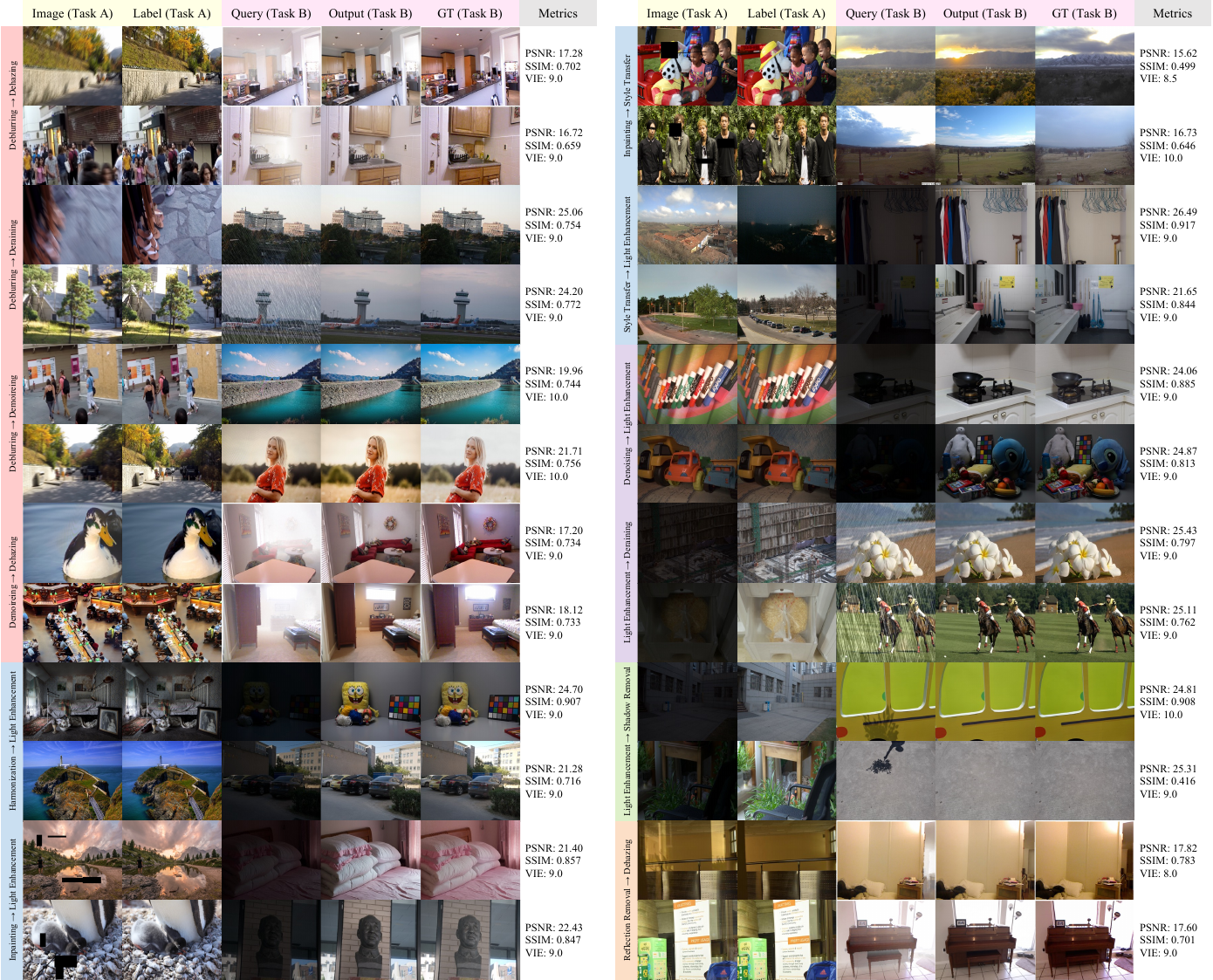}
    \caption{Representative Examples of Cross-Task In-Context Learning on Twelve Pairs with Gemini}
    \label{fig:fig4}
\vspace{-8pt}
\end{figure*}

Under the best-of-10 protocol, Table~\ref{tab:3} presents quantitative comparisons across twelve representative task pairs on Gemini and Seedream, each compared with its fixed-prompt baseline.
Figure~\ref{fig:fig4} illustrates representative examples of cross-task ICL on Gemini, demonstrating how the VLM adapts flexibly to diverse tasks when conditioned on implicit text prompts. Among these tasks, the proposed implicit prompt outperforms the fixed-prompt baseline on at least two of the three evaluation metrics for both models, and the third metric remains very close. Notably, VIEScore improves throughout, highlighting the effectiveness of our implicit prompt-driven generalization mechanism. Additional relevant Seedream examples are provided in Figure~\ref{fig:fig6}, showing that the qualitative gains of T2T-VICL are not specific to Gemini.

\subsection{Cross-Task Diversity and Adaptability}
T2T-VICL demonstrates consistent gains across diverse low-level vision tasks such as dehazing, deraining, and demoireing. Moreover, the same prompting mechanism seamlessly extends to generation-oriented tasks such as low-light enhancement, effectively maintaining spatial layout and identity fidelity. Unlike prior approaches that are often confined to semantically similar domains, our method adapts seamlessly across both intra-category tasks (e.g., harmonization vs. light enhancement, demoireing vs. dehazing) and inter-category tasks (e.g., reflection removal vs. dehazing). These findings highlight the scalability and flexibility of our approach, showing that T2T-VICL can naturally bridge both perceptually related and unrelated tasks under a single model.

\subsection{Generalization to Distant Tasks} Beyond absolute numbers, an important property is cross-task consistency. This consistency underlines the advantage of framing all tasks under a unified in-context prompting mechanism. Moreover, T2T-VICL generalizes to semantically distant task pairs such as denoising versus light enhancement, where task-specialized models tend to degrade without task-specific retraining. This highlights robustness as a key byproduct of the unified in-context paradigm.

\subsection{Cross-Task Results in Second-Tier Group}
Across another nine cross-task pairs in Table~\ref{tab:4}, our method consistently improves VIEScore over the fixed-prompt baseline on both Gemini and Seedream. The other two IQA metrics have slight declines in several cases. Interestingly, for these compositions, improvements are primarily reflected in the VIEScore rather than in the traditional fidelity metrics.
Since VIEScore assesses instruction following and perceptual quality, it is more sensitive to semantic preservation than to strict pixel-level alignment. In most cases, the visual consistency of the generated results remains comparable to that of the top-tier group, as shown in Figure~\ref{fig:fig5}. See detailed discussion in Appendix Sec.~\ref{sec:viescore}.

\begin{table*}[ht]
\centering
\scriptsize
\setlength{\tabcolsep}{0.5pt}
\renewcommand{\arraystretch}{0.96}
\captionsetup{width=0.98\textwidth, skip=4pt, type=table}
\caption{Average of ground-truth-based best-of-10 evaluation results in the second-tier group. Generated by Gemini and Seedream with Qwen prompts, compared to the fixed-prompt baseline.}
\vspace{-1pt}
\label{tab:4}

\begin{adjustbox}{width=\textwidth}
\begin{tabular}{>{\raggedright\arraybackslash\leftskip=2pt}m{\dimexpr4.09cm+3pt\relax} *{5}{>{\centering\arraybackslash}m{0.8cm}} >{\centering\arraybackslash}m{1cm} *{5}{>{\centering\arraybackslash}m{0.8cm}} >{\centering\arraybackslash}m{1cm}}
\toprule
\multirow{3}{*}{\textbf{Task}}
& \multicolumn{6}{c}{\textbf{Gemini 2.5 Flash}}
& \multicolumn{6}{c}{\textbf{Seedream 4.0}} \\
& \multicolumn{2}{c}{\textbf{PSNR}$\uparrow$}
& \multicolumn{2}{c}{\textbf{SSIM}$\uparrow$}
& \multicolumn{2}{c}{\textbf{VIEScore}$\uparrow$}
& \multicolumn{2}{c}{\textbf{PSNR}$\uparrow$}
& \multicolumn{2}{c}{\textbf{SSIM}$\uparrow$}
& \multicolumn{2}{c}{\textbf{VIEScore}$\uparrow$} \\
& \textcolor{gray}{\textbf{Fixed}} & \textcolor{gray}{\textbf{Ours}}
& \textcolor{gray}{\textbf{Fixed}} & \textcolor{gray}{\textbf{Ours}}
& \textcolor{gray}{\textbf{Fixed}} & \textcolor{gray}{\textbf{Ours}}
& \textcolor{gray}{\textbf{Fixed}} & \textcolor{gray}{\textbf{Ours}}
& \textcolor{gray}{\textbf{Fixed}} & \textcolor{gray}{\textbf{Ours}}
& \textcolor{gray}{\textbf{Fixed}} & \textcolor{gray}{\textbf{Ours}} \\
\midrule

\rowcolor{lightred!30}
Dehazing $\rightarrow$ Deraining
& \cellcolor{lightred!60}19.09 & 19.01
& \cellcolor{lightred!60}0.518 & 0.505
& 6.39 & \cellcolor{lightred!60}\tabgainsubscript{7.31}{+0.92}
& 14.47 & \cellcolor{lightred!60}14.49
& 0.385 & \cellcolor{lightred!60}0.386
& 6.41 & \cellcolor{lightred!60}\tabgainsubscript{6.76}{+0.35} \\
\grayhline

\rowcolor{lightred!30}
Deraining $\rightarrow$ Demoireing
& \cellcolor{lightred!60}15.29 & 15.00
& \cellcolor{lightred!60}0.583 & 0.580
& 5.83 & \cellcolor{lightred!60}\tabgainsubscript{7.24}{+1.41}
& \cellcolor{lightred!60}11.84 & 10.95
& \cellcolor{lightred!60}0.451 & 0.418
& 7.38 & \cellcolor{lightred!60}\tabgainsubscript{7.40}{+0.02} \\
\grayhline

\rowcolor{lightblue!30}
Colorization $\rightarrow$ Style Transfer
& \cellcolor{lightblue!60}13.35 & 13.03
& \cellcolor{lightblue!60}0.465 & 0.434
& 5.10 & \cellcolor{lightblue!60}\tabgainsubscript{7.37}{+2.27}
& \cellcolor{lightblue!60}9.91 & 9.54
& \cellcolor{lightblue!60}0.333 & 0.314
& 7.99 & \cellcolor{lightblue!60}\tabgainsubscript{8.09}{+0.10} \\
\grayhline

\rowcolor{lightblue!30}
Harmonization $\rightarrow$ Style Transfer
& \cellcolor{lightblue!60}13.32 & 13.14
& \cellcolor{lightblue!60}0.465 & 0.458
& 4.79 & \cellcolor{lightblue!60}\tabgainsubscript{5.66}{+0.87}
& \cellcolor{lightblue!60}10.20 & 9.63
& \cellcolor{lightblue!60}0.354 & 0.331
& 7.32 & \cellcolor{lightblue!60}\tabgainsubscript{7.62}{+0.30} \\
\grayhline

\rowcolor{lightblue!30}
Inpainting $\rightarrow$ Colorization
& \cellcolor{lightblue!60}21.31 & 20.75
& \cellcolor{lightblue!60}0.843 & 0.742
& 5.23 & \cellcolor{lightblue!60}\tabgainsubscript{7.39}{+2.16}
& \cellcolor{lightblue!60}13.53 & 12.08
& \cellcolor{lightblue!60}0.447 & 0.397
& 5.86 & \cellcolor{lightblue!60}\tabgainsubscript{6.96}{+1.10} \\
\grayhline

\rowcolor{lightblue!30}
Light Enhancement $\rightarrow$ Colorization
& \cellcolor{lightblue!60}22.72 & 21.04
& \cellcolor{lightblue!60}0.860 & 0.753
& 5.05 & \cellcolor{lightblue!60}\tabgainsubscript{6.86}{+1.81}
& 11.01 & \cellcolor{lightblue!60}12.09
& \cellcolor{lightblue!60}0.436 & 0.422
& 4.59 & \cellcolor{lightblue!60}\tabgainsubscript{7.28}{+2.69} \\
\grayhline

\rowcolor{lightyellow!30}
Shadow Removal $\rightarrow$ Reflection Removal
& \cellcolor{lightyellow!50}21.79 & 21.44
& \cellcolor{lightyellow!50}0.761 & 0.739
& 6.24 & \cellcolor{lightyellow!50}\tabgainsubscript{6.34}{+0.10}
& \cellcolor{lightyellow!50}14.96 & 13.10
& \cellcolor{lightyellow!50}0.532 & 0.459
& 7.13 & \cellcolor{lightyellow!50}\tabgainsubscript{7.17}{+0.04} \\
\grayhline

\rowcolor{lightpurple!30}
Deraining $\rightarrow$ Style Transfer
& \cellcolor{lightpurple!60}13.31 & 13.10
& 0.430 & \cellcolor{lightpurple!60}0.447
& 4.67 & \cellcolor{lightpurple!60}\tabgainsubscript{5.66}{+0.99}
& \cellcolor{lightpurple!60}11.02 & 9.76
& \cellcolor{lightpurple!60}0.375 & 0.318
& 6.66 & \cellcolor{lightpurple!60}\tabgainsubscript{7.33}{+0.67} \\
\grayhline

\rowcolor{lightorange!30}
Shadow Removal $\rightarrow$ Deraining
& \cellcolor{lightorange!50}19.49 & 18.98
& \cellcolor{lightorange!50}0.555 & 0.538
& 7.54 & \cellcolor{lightorange!50}\tabgainsubscript{8.20}{+0.66}
& \cellcolor{lightorange!50}17.37 & 15.25
& \cellcolor{lightorange!50}0.487 & 0.431
& 7.51 & \cellcolor{lightorange!50}\tabgainsubscript{7.68}{+0.17} \\
\bottomrule
\end{tabular}
\end{adjustbox}
\vspace{-3pt}
\end{table*}

\begin{figure*}[ht]
    \centering
    \includegraphics[width=\textwidth]{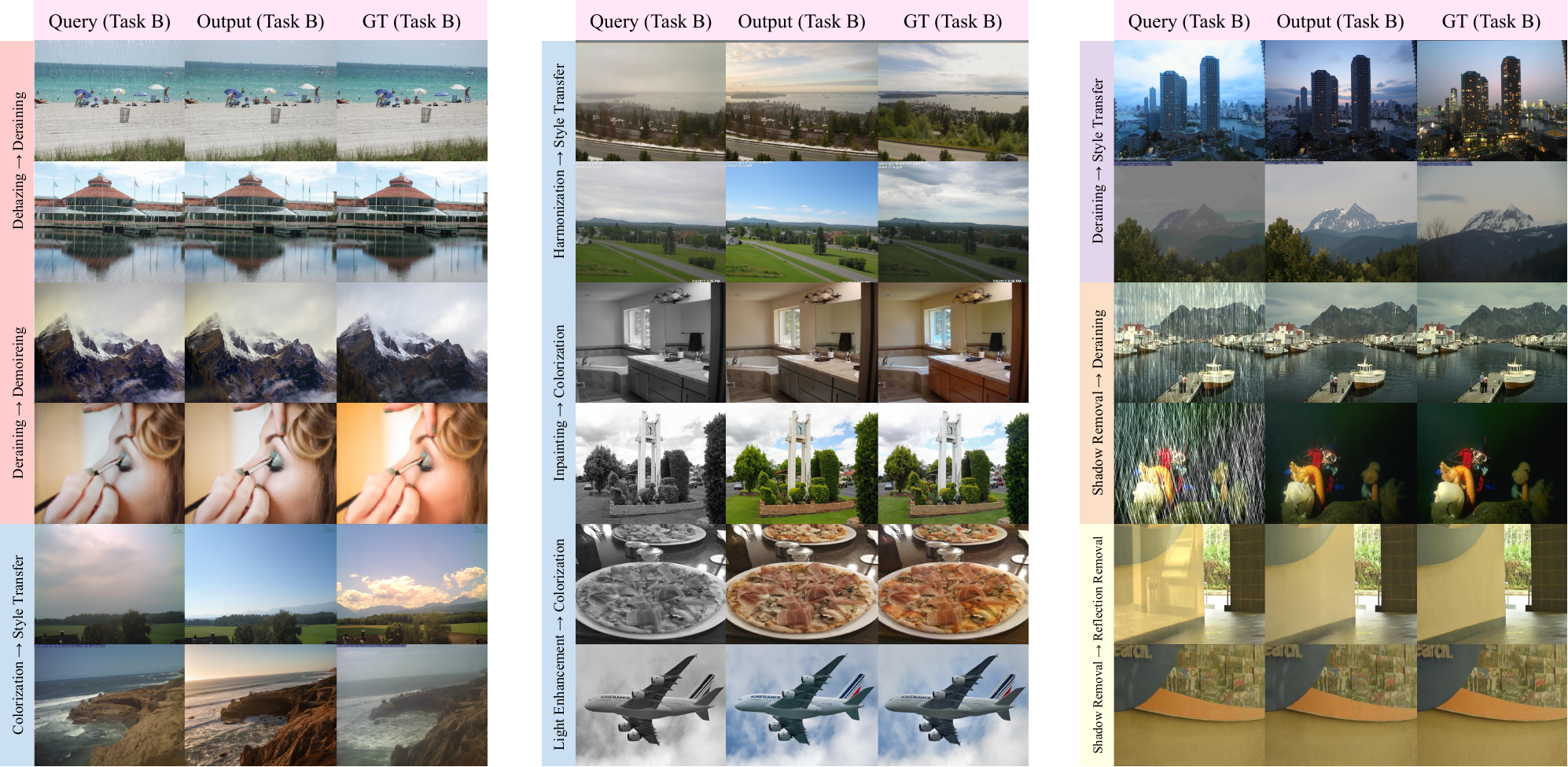}
    \caption{Representative Examples of Cross-Task In-Context Learning on Nine Pairs with Gemini}
    \label{fig:fig5}
\vspace{-8pt}
\end{figure*}

\subsection{Comparison with Existing Methods}

\noindent \textbf{Comparison protocol.} T2T-VICL performs cross-task inference from a Task-$A$ demonstration to a Task-$B$ query. For each target task, all competitor methods are evaluated on 100 input/output pairs drawn from the same Task-$B$ dataset. When several Task $A$ and $B$ pairs share the same Task $B$, we report results relevant to one Task $A$ for T2T-VICL. Each model uses its default inference settings and performs one generation per query. Accordingly, T2T-VICL also uses its single-shot outputs here to align with the competitors.

Table~\ref{tab:competitor} compares our T2T-VICL with MAE-VQGAN~\cite{bar2022visual}, InstructDiffusion~\cite{geng2024instructdiffusion}, and VisualCloze~\cite{li2025visualcloze}. The inpainting task is excluded from the comparison, as it was not used as a Task $B$ in our main cross-task evaluation.
Same-task VICL methods receive a same-task Task-$B$ demonstration which is distinct from the query, whereas InstructDiffusion receives only the Task-$B$ query and task instruction. All competitors therefore receive direct target task evidence as context, while our method must infer from a demonstration of a different task.
Despite the harder cross-task setting and the absence of a direct demonstration of the target task, T2T-VICL still obtains the highest PSNR and SSIM on several target tasks and performs at a comparable level overall, providing evidence that cross-task transfer is effective.
Figure~\ref{fig:fig7} in Appendix Sec.~\ref{sec:example} provides the corresponding qualitative comparison across the target tasks.

\begin{table}[t]
\centering
\captionsetup{width=0.99\columnwidth, skip=4pt, type=table}
\caption{Per-target PSNR and SSIM comparison. For each target task, competitors are evaluated on 100 independent queries from the same Task-$B$ data as T2T-VICL. The method $^{*}$ does not support the Style Transfer task.}
\label{tab:competitor}
\scriptsize
\setlength{\tabcolsep}{2.3pt}
\renewcommand{\arraystretch}{1.05}
\resizebox{\columnwidth}{!}{%
\begin{tabular}{l*{4}{cc}}
\toprule
\multirow{3}{*}{\textbf{Target Task}} & \multicolumn{2}{c}{\textbf{T2T-VICL}} & \multicolumn{2}{c}{\textbf{MAE-VQGAN}} & \multicolumn{2}{c}{\textbf{VisualCloze}} & \multicolumn{2}{c}{\textbf{InstructDiffusion}$^{*}$} \\
& \multicolumn{2}{c}{\textit{Cross-task VICL}} & \multicolumn{4}{c}{\textit{Same-task VICL}} & \multicolumn{2}{c}{\textit{Query + Instruction}} \\
\cmidrule(lr){2-3} \cmidrule(lr){4-7} \cmidrule(lr){8-9}
& \textbf{PSNR}$\uparrow$ & \textbf{SSIM}$\uparrow$ & \textbf{PSNR}$\uparrow$ & \textbf{SSIM}$\uparrow$ & \textbf{PSNR}$\uparrow$ & \textbf{SSIM}$\uparrow$ & \textbf{PSNR}$\uparrow$ & \textbf{SSIM}$\uparrow$ \\
\midrule
\rowcolor{lightred!30} Deblurring          & \underline{19.47} & \underline{0.592} & 14.73 & 0.451 & \textbf{20.20} & \textbf{0.608} & 14.07 & 0.410 \\
\rowcolor{lightred!30} Dehazing            & \underline{11.44} & \underline{0.461} & 10.20 & 0.337 & 9.56 & 0.276 & \textbf{12.75} & \textbf{0.593} \\
\rowcolor{lightred!30} Demoireing          & \textbf{15.26} & 0.566 & 11.97 & \underline{0.611} & 10.43 & 0.449 & \underline{14.40} & \textbf{0.620} \\
\rowcolor{lightred!30} Denoising           & 18.59 & 0.661 & 20.41 & \textbf{0.729} & \textbf{21.68} & 0.682 & \underline{20.49} & \underline{0.700} \\
\rowcolor{lightred!30} Deraining           & \textbf{18.77} & \textbf{0.509} & 15.43 & 0.360 & \underline{16.85} & \underline{0.453} & 12.98 & 0.356 \\
\rowcolor{lightyellow!30} Reflection Removal & \textbf{19.42} & \textbf{0.660} & \underline{17.10} & \underline{0.549} & 16.49 & 0.539 & 14.37 & 0.530 \\
\rowcolor{lightyellow!30} Shadow Removal     & \underline{16.56} & 0.246 & 15.03 & \underline{0.282} & 11.17 & 0.194 & \textbf{17.79} & \textbf{0.441} \\
\rowcolor{lightblue!30} Colorization       & \textbf{20.08} & \textbf{0.662} & \underline{14.78} & 0.350 & 11.24 & 0.326 & 11.65 & \underline{0.395} \\
\rowcolor{lightblue!30} Harmonization      & \textbf{17.02} & \textbf{0.483} & \underline{14.79} & \underline{0.386} & 13.88 & 0.380 & 12.13 & 0.352 \\
\rowcolor{lightblue!30} Light Enhancement  & 10.33 & \underline{0.421} & \underline{13.15} & 0.406 & 9.88 & 0.287 & \textbf{16.09} & \textbf{0.529} \\
\rowcolor{lightblue!30} Style Transfer     & \textbf{13.05} & \textbf{0.441} & \underline{11.06} & \underline{0.424} & 10.79 & 0.373 & -- & -- \\
\bottomrule
\end{tabular}%
}
\vspace{-12pt}
\end{table}

\subsection{Discussion}
Overall, our framework demonstrates strong cross-task generalization and robust visual consistency across a wide range of low-level vision tasks. By leveraging implicit prompt representations, the model effectively adapts demonstrations to unseen queries without task-specific fine-tuning, achieving top-tier VIEScores in many cases. However, due to the inherent complexity of cross-task compositions, a few cases with limited reference samples occasionally produce content that does not exist in the original images, slightly lowering the fidelity-based indicators. Despite these imperfections, the overall visual consistency remains close to that of the top group. Future work will focus on adaptive balancing between semantic-level and pixel-level objectives to further improve realism and generalization across diverse conditions.

\paragraph{Limitations.}
Since our scope is primarily on low-level vision tasks, we select 12 representative low-level tasks, covering most commonly studied low-level tasks in the literature and exceeding the task coverage of many VICL and prompt-enhanced works in this field (e.g., MAE-VQGAN~\cite{bar2022visual}, Painter~\cite{wang2023images}, ProRes~\cite{ma2023prores}, PromptIR~\cite{potlapalli2023promptir}, InstructIR~\cite{conde2024instructir}). Notably, the number of cross-task experiments grows quadratically with the number of tasks, compared to same-task VICL, and we will consider more niche tasks in the future. In some representative higher-level tasks, such as geometric problem-solving and spatial reasoning, VLMs may exhibit weak baseline performance, thus the practical applicability of our approach in these scenarios still requires further investigation. Meanwhile, task transfer capability in VLMs is not unlimited and might depend on the underlying similarity between low-level tasks. Our results indicate that implicit task transfer is feasible but also constrained, which is partially consistent with previous studies such as Taskonomy~\cite{zamir2018taskonomy}. The teacher may also transfer its phrasing and failure modes to the student.

\section{Conclusion}
\label{sec:concl}

In conclusion, we propose T2T-VICL, the first cross-task visual in-context learning pipeline that enables collaboration among multiple VLMs, together with an automatic reasoning mechanism. Our framework systematically explores the largely uncharted boundaries of cross-task transfer in VICL. This collaborative paradigm fills a gap in understanding how heterogeneous VLMs can jointly reason and adapt across tasks, and we anticipate that our findings will encourage further investigation into the mechanisms that underpin cross-task transfer and the broader generalization capacity of multimodal models.


\clearpage
\bibliography{ref}

\appendix
\clearpage

\section*{Appendix}
\label{sec:appendix}

\section{Dataset Details}
\label{sec:data}

We build our T2T-VICL benchmark by assembling representative paired datasets across diverse low-level image processing tasks. For each task, we use publicly released datasets without altering their ground-truth semantics when available. Where appropriate, we apply lightweight, clearly documented preprocessing so that training and evaluation are consistent across tasks. The dataset selection and provenance follow Sec.~\ref{sec:setup} in the main paper. The following subsections detail the origin of each dataset, the typical resolution, and the data split employed in our experiments.

\subsection{Per-Dataset Descriptions}
\noindent \textbf{GOPRO (deblurring).} We use this large motion deblurring dataset introduced for dynamic scene deblurring. It contains 3,214 blurry and sharp pairs captured from high-speed video and typically provided at 1280 $\times$ 720 resolution. This dataset provides realistic, non-uniform motion blur that matches the motion-blur restoration scenario in our cross-task experiments.

\noindent \textbf{D-HAZY (dehazing).} D-HAZY is a synthetic dehazing dataset produced by applying the Koschmieder atmospheric scattering model to images with depth maps from the Middlebury and NYU Depth datasets. The dataset contains over 1,400 hazy and haze-free image pairs and is widely used to benchmark single-image dehazing approaches. We use the official image pairs and follow the standard splits reported by the dataset authors.

\noindent \textbf{UHDM (demoireing).} We use this ultra-high-definition demoireing dataset, which is composed of 5,000 real 4K image pairs collected for moire removal and scale-robust evaluation. UHDM is appropriate for demoireing examples in our benchmark because it contains diverse real-world moire patterns at 4K scale. During training, we downsample images to match the target model resolutions while retaining high-resolution examples for qualitative analysis.

\noindent \textbf{SIDD (denoising).} The Smartphone Image Denoising Dataset, denoted as SIDD, provides real noisy smartphone photographs and matching ground truth. The dataset contains approximately 30,000 noisy images captured across multiple scenes and devices, serving as a standard benchmark for real-world image denoising. We use the official training and benchmark splits for our denoising experiments.

\noindent \textbf{Rain13K (deraining).} We use the synthetic rainy and clear pairs assembled by \citet{jiang2020multi} and subsequently adopted by MPRNet~\cite{zamir2021multi}. Commonly referred to as Rain13K, this collection aggregates 11,200 training pairs from Rain14000, 1,800 from Rain1800, 700 from the training portion of Rain800, and 12 from Rain12. The rainy images serve as inputs and their aligned clean counterparts as targets.

\noindent \textbf{iHarmony4 (harmonization).} iHarmony4 is a large image harmonization benchmark composed of four sub-datasets, comprising HCOCO, HAdobe5k, HFlickr, and Hday2night. Together, these sub-datasets provide tens of thousands of composed images and corresponding real images. We pick the Hday2night set for harmonization experiments.

\noindent \textbf{DIV2K (inpainting).} DIV2K contains 1,000 high-quality images with diverse content at 2K resolution. For inpainting, our source pool is the 800-image DIV2K training split. We apply synthetic black masks to generate the inputs and retain the corresponding clean images as targets.

\noindent \textbf{ImageNet (colorization).} Following the data construction process in Colorful Image Colorization~\cite{deng2009imagenet,zhang2016colorful}, we utilize a modified subset constructed from 5,000 ImageNet color images. Grayscale conversion serves as the input while the original RGB version serves as the target. Both counterparts are stored at 400 $\times$ 400 resolution.

\noindent \textbf{LoL (light enhancement).} The Low-Light dataset, referred to as LoL, provides paired low-light and normal-light images. The most commonly used version LoL-v1 contains 500 image pairs. We adopt the official image pairs in our light-enhancement experiments.

\noindent \textbf{SIR$^2$ (reflection removal).} The SIR$^2$ benchmark provides diverse images from both controlled indoor and real-world outdoor scenes, each with ground truth of background and reflection. We use the SIR$^2$ ground-truth pairs for our reflection removal experiments.

\noindent \textbf{ISTD (shadow removal).} The Image Shadow Triplets dataset, known as ISTD, contains 1,870 pairs of shadow and shadow-free images. We use the ISTD pairs for shadow removal evaluations.

\noindent \textbf{Night2Day (style transfer).} For style transfer, we use the publicly available Night2Day dataset, which includes images of outdoor scenes under various environmental appearances such as snow, autumn, dusk, and fog. We manually pair images representing the same underlying scene under different visual conditions to construct the image pairs required for our experiments.

\paragraph{Data splits and preprocessing.} For all datasets mentioned above, we use official splits when available and otherwise adopt the 70/30 split described in Sec.~\ref{sec:setup}. We keep our data processing protocols, particularly for synthetic variants such as the inpainting data constructed from DIV2K and generated with synthetic black masks.

\subsection{Pair Construction for Cross-Task}
For every source dataset representing Task $A$, we sample $N$ examples and pair them with $N$ examples drawn from the target dataset representing Task $B$ to form cross-task demonstration pairs. To maximize the diversity of implicit descriptions used to teach the student VLM, we generate candidate teacher texts across various source and target task pairs, and then we apply embedding-based clustering and deduplication to retain high-variance and information-rich implicit prompts.

\section{Prompt Design}
\label{sec:prompt}

Two different conditioning strategies are compared in our experiments: (1) a fixed template prompt that merely describes the input and output demonstration structure, and (2) implicit prompts generated at inference time by the fine-tuned Qwen student model. The implicit prompts are intended to describe visual differences and transformation goals between the provided Task $A$ input and output pair and the Task $B$ query input without explicitly naming the tasks. The implicit prompts therefore act as a content-dependent, transformation-specific instruction that guides the downstream VLM to adapt the example transformation to a new input. This design is central to the capability of T2T-VICL in performing cross-task visual in-context learning.

\subsection{Fixed Prompt}
We used the following prompt as the baseline in our fixed-prompt evaluation experiments:

\begin{tcolorbox}[colback=gray!20, colframe=gray!20, boxrule=0pt, left=3pt, right=3pt, top=3pt, bottom=3pt]
\emph{This is a visual in-context learning task. The first two images are an input and output of Task A. The third image is the input for Task B. The goal is to perform Task B on the third image and generate an output image, learning from Task A.}
\end{tcolorbox}

\noindent This fixed prompt provides clear structural cues but does not communicate task-specific differences, for example, whether the change involves removing linear occlusions or suppressing uniform noise. Therefore, it acts as a conservative baseline for evaluating the benefits of descriptive implicit instructions.

\noindent Similarly, the fixed prompt applied in the same-task VICL experiment is as follows:

\begin{tcolorbox}[colback=gray!20, colframe=gray!20, boxrule=0pt, left=3pt, right=3pt, top=3pt, bottom=3pt]
\emph{This is a visual in-context learning task. The first two images are an input and output of an image processing task. The third image is the input for the SAME task. The goal is to perform this task on the third image and generate an output image, learning from the pair before.}
\end{tcolorbox}

\subsection{Implicit Prompt (Qwen)}
\textbf{How it is created.} For each sampled image tuple $\mathcal{X}_t=(I_A^{in}, I_A^{l}, I_B^{in}, I_B^l)$ in Eq.~\ref{eq:teacher-input}, a large teacher VLM is prompted with four images and an open-ended instruction, requesting a comparison of the observed effects in these images and generation of a descriptive instruction that would guide another model to transform $I_B^{in}$ into $I_B^{out}$. This process elicits a rich, descriptive text. The candidate texts undergo embedding clustering, deduplication, and filtering to keep high-variance, informative implicit prompts which then serve as training targets for our small student VLM to learn to generate such implicit prompts given similar image tuples $\mathcal{X}_s=(I_A^{in}, I_A^{l}, I_B^{in})$.

\noindent \textbf{What the implicit prompt contains.} Typical implicit prompts contain three key components: (1) target goals regarding desired visual attributes in the output, (2) descriptions of input degradation to be mitigated, and (3) instructions on modifying visual statistics such as preserving texture, restoring contrast, and avoiding color casts. Importantly, the generated language avoids revealing explicit task-specific terms (e.g., deraining, denoising, and deblurring), forcing the model to rely on descriptive visual cues and allowing cross-task transformation transfer.

\noindent \textbf{Why it helps.} Because the implicit prompt encodes the target visual changes in fine-grained, task-agnostic language, it provides the downstream generator with actionable, content-sensitive guidance. This guidance improves adaptation when the transformation required by Task $B$ shares similar underlying transformation principles with Task $A$, despite their differences in surface form.

\noindent \textbf{Example.}
Below we give one long, representative teacher output used as an illustrative example in the supplementary. This example is intentionally verbose to demonstrate the kinds of distinctions the implicit prompt encodes. A single long instruction like this is used as a training target for the student VLM.

\begin{tcolorbox}[colback=gray!20, colframe=gray!20, boxrule=0pt, left=3pt, right=3pt, top=3pt, bottom=3pt]

\emph{This is a visual in-context learning task. The first two images are an input and output of Task A. The third image is the input for Task B. The goal is to perform Task B on the third image and generate an output image, learning from Task A.}
\vspace{1pt}

\emph{\textbf{Analysis of Tasks A and B}}

\emph{\textbf{Input Descriptions}}

\begin{itemize}

\item \emph{\textbf{Task A:} The input image shows a dimly lit room with stacked boxes containing various items, including a basketball. The lighting is low, making details difficult to discern.}

\item \emph{\textbf{Task B:} The input image is a black-and-white photograph of a person wearing sunglasses and a headscarf, holding a phone to their ear. The image lacks color and has a monochromatic tone.}

\end{itemize}

\emph{\textbf{Visual Changes}}

\begin{itemize}

\item \emph{\textbf{Task A:} The output image is brighter, with improved lighting that enhances visibility and detail of the objects within the room.}

\item \emph{\textbf{Task B:} The output image is in color, transforming the black-and-white photo into a full-color representation, adding vibrancy and depth.}

\end{itemize}

\emph{\textbf{Task Differences}}

\begin{itemize}

\item \emph{\textbf{Task A:} Task A focuses on enhancing brightness and visibility, likely through illumination or contrast adjustments.}

\item \emph{\textbf{Task B:} Task B emphasizes color restoration, converting a grayscale image into a colorful one, which alters the overall aesthetic and provides more visual information.}

\end{itemize}

\end{tcolorbox}

\noindent Similarly, one example of an implicitly enhanced prompt in the same-task study is shown below. Same-task VICL does not include the \emph{Task Differences} item.


\begin{tcolorbox}[colback=gray!20, colframe=gray!20, boxrule=0pt, left=3pt, right=3pt, top=3pt, bottom=3pt]

\emph{This is a visual in-context learning task. The first two images are an input and output of an image processing task. The third image is the input for the same task. The goal is to perform this task on the third image and generate an output image, learning from the pair before.}

\end{tcolorbox}

\begin{tcolorbox}[colback=gray!20, colframe=gray!20, boxrule=0pt, left=3pt, right=3pt, top=3pt, bottom=3pt]

\emph{\textbf{Image Descriptions}}

\emph{Picture 1 shows a monochrome shot of a racer on a motorcycle, while Picture 2 presents the same scene with vibrant, saturated colors restored.}

\emph{\textbf{Visual Changes}}

\emph{The task transforms the grayscale representation into a vivid, full-color depiction, enhancing the visual richness and detail of the subject.}

\emph{The third image, a grayscale kite on grass, would similarly gain vivid color saturation to match the enhanced palette of the first two images.}

\end{tcolorbox}

\section{Inference Details}
\label{sec:infer}

\begin{figure*}[t]
    \centering
    \includegraphics[width=0.9\textwidth]{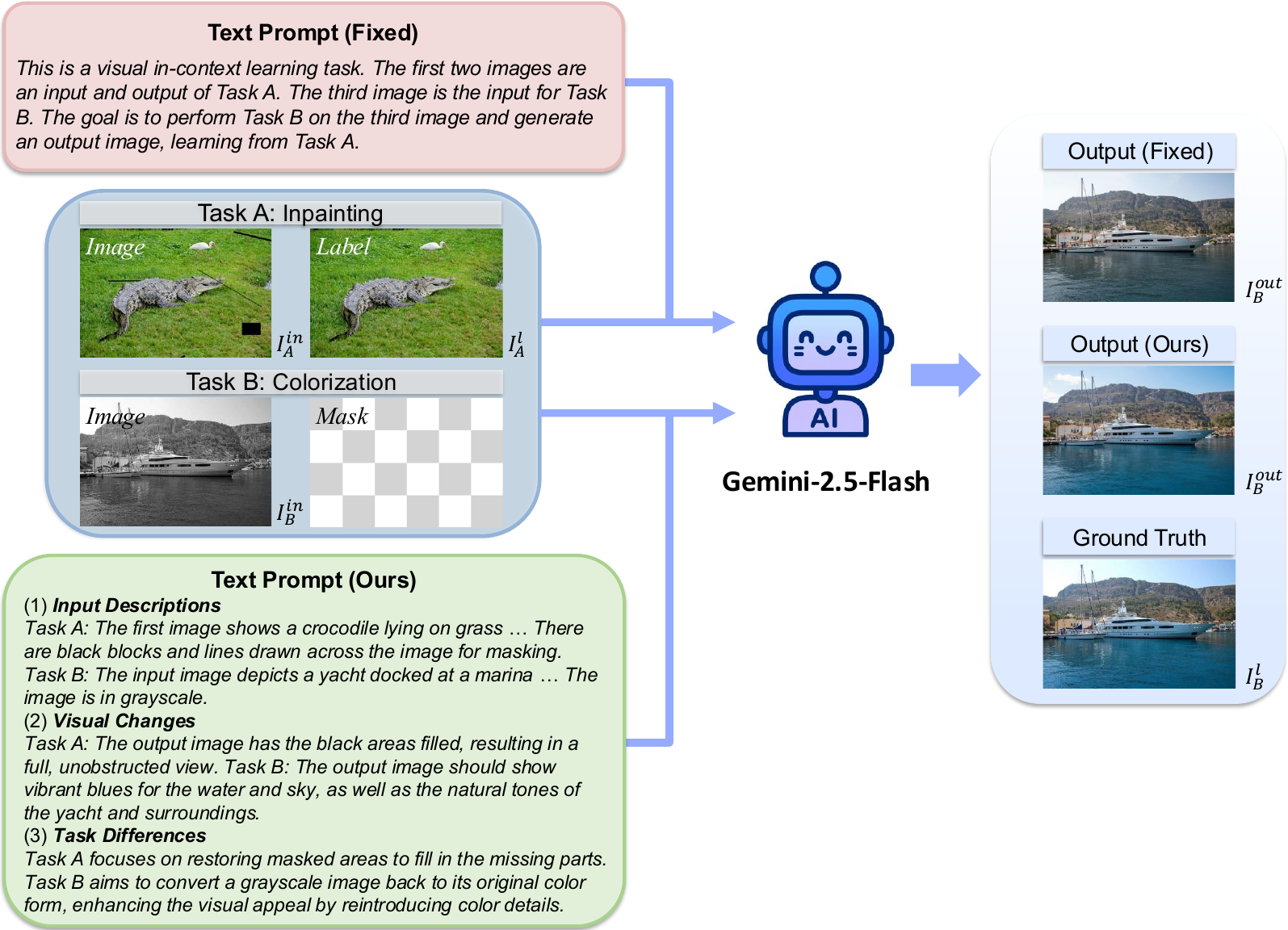}
    \caption{Detailed Example of Cross-Task Visual In-Context Learning in Inference Stage}
    \label{fig:fig3}
\vspace{-8pt}
\end{figure*}

Figure~\ref{fig:fig3} gives a concrete inference-stage example to clarify how the fixed prompt and the implicit prompt are used. Both settings receive identical visual inputs, consisting of an input-output demonstration pair from task $A$ and a query input from task $B$.
T2T-VICL first leverages the trained student VLM to produce an implicit prompt from the same images, and the generated prompt is then concatenated with the fixed structural instruction and passed to the frozen VLM along with the images.
The figure visualizes this complete route from visual prompt construction to final output, showing how the implicit prompt serves as an interpretable bridge between a task-$A$ demonstration and a task-$B$ query.

\section{Additional Results and Analyses}

\subsection{Additional Qualitative Examples}
\label{sec:example}

\begin{figure*}[t]
    \centering
    \includegraphics[width=\textwidth]{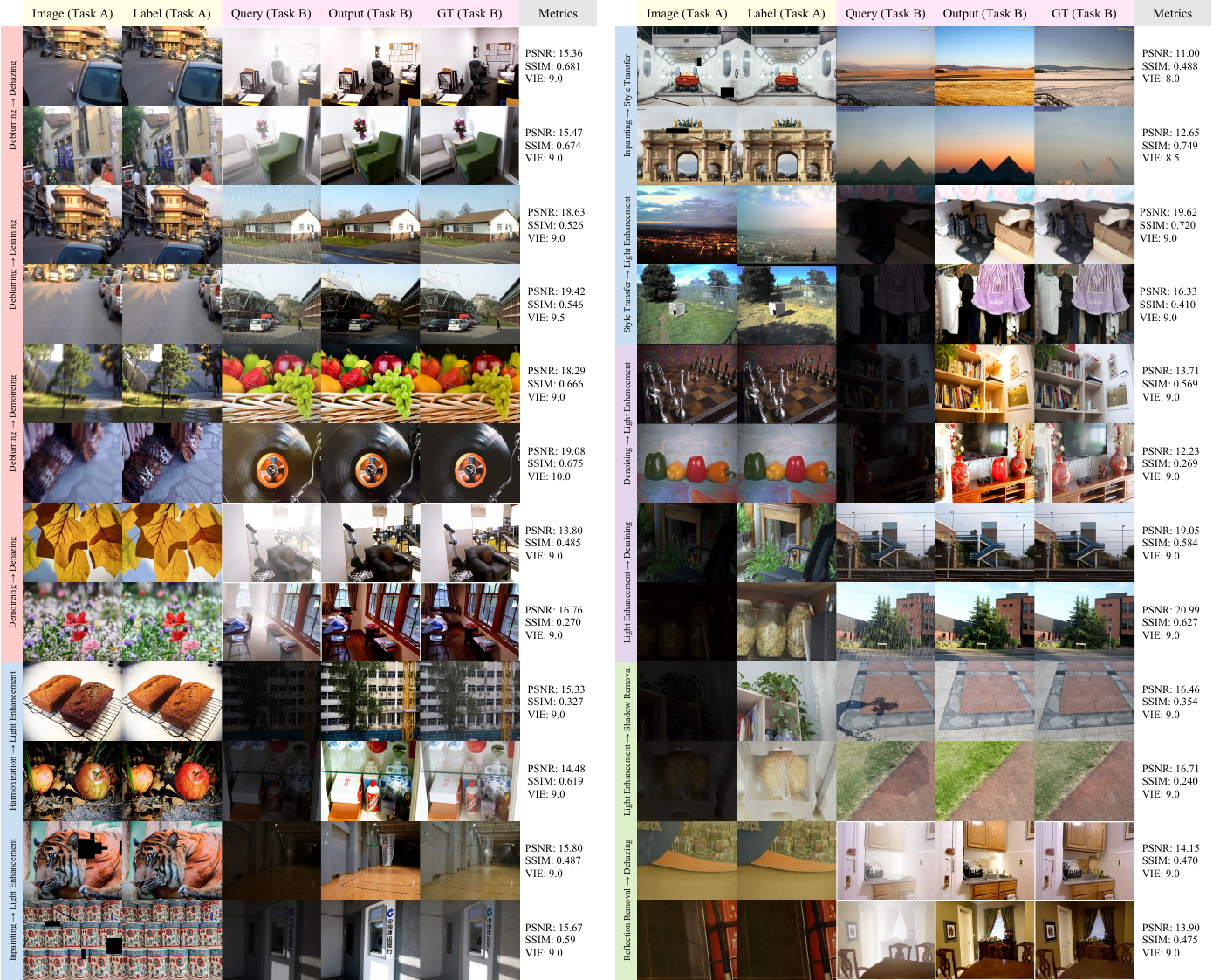}
    \caption{Representative Examples of Cross-Task In-Context Learning on Twelve Pairs with Seedream}
    \label{fig:fig6}
\vspace{-8pt}
\end{figure*}

\begin{figure*}[ht]
    \centering
    \includegraphics[width=\textwidth]{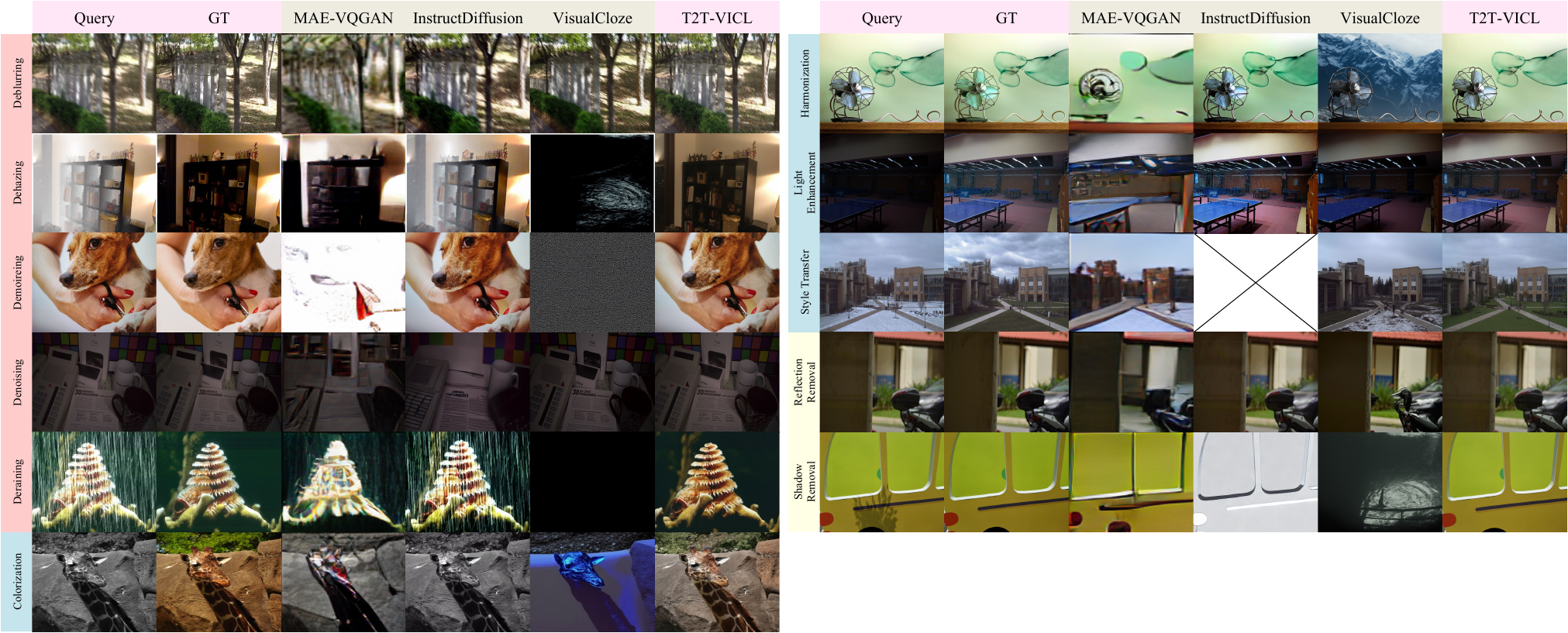}
    \caption{Qualitative Comparison of Generated Results Between Existing Methods and Ours}
    \label{fig:fig7}
\vspace{-8pt}
\end{figure*}

The Seedream qualitative examples in Figure~\ref{fig:fig6} complement the best-of-10 results in the main paper.
Seedream and Gemini experiments use the same cross-task inference protocol, where the fixed prompt and the Qwen-generated implicit prompt are applied to identical task-$A$ demonstrations and task-$B$ queries, varying only the final image-editing backbone.
Additionally, Figure~\ref{fig:fig7} supplements the comparison results in Table~\ref{tab:competitor} with representative outputs from T2T-VICL and competing methods across the evaluated target tasks.

\subsection{Second-Tier Typed Result Explanation}
\label{sec:viescore}

In the best-of-10 second-tier group, T2T-VICL consistently improves VIEScore despite occasional decreases in PSNR or SSIM. These cases may not necessarily be interpreted as failures of cross-task VICL, as qualitative evidence in Figure~\ref{fig:fig5} shows that the visual consistency of these generated outputs remains comparable to that of the top-tier group.

The gap instead might indicate a mismatch between pixel-aligned fidelity metrics and the goal of several specific target tasks, especially when the target task is closer to conditional generation than deterministic restoration. This issue is particularly evident in colorization and style transfer tasks, which inherently possess one-to-many mapping characteristics where multiple plausible outputs can correspond to a single input~\cite{su2020instance,kang2023ddcolor}.

In colorization, an output that remains close to the grayscale input can preserve luminance structure and obtain competitive PSNR and SSIM, even if it fails to add plausible chroma. Recent colorization studies still commonly report PSNR and SSIM, but they increasingly supplement them with perceptual, distributional, or color-statistical measures such as LPIPS, FID, and colorfulness-related metrics~\cite{wu2021towards,kang2023ddcolor}. This is not because PSNR and SSIM are invalid metrics, but because pixel-level fidelity alone cannot fully capture color plausibility under a one-to-many target distribution. A VIEScore improvement with a slight PSNR or SSIM drop can therefore still correspond to better task performance.

Style transfer operates under an even weaker one-to-one correspondence assumption, where successfully stylized images must preserve semantic content while intentionally changing appearance, texture, and color statistics. Consequently, style-transfer literature often prioritizes perceptual measures and human preference evaluations over traditional PSNR or SSIM metrics~\cite{deng2022stytr2, wright2022artfid}.

These related observations actually support our use of VIEScore as the primary signal for the second-tier group, since VIEScore is able to evaluate whether the output follows the intended visual instruction and preserves the relevant content. The consistent VIEScore gains alongside the mixed pixel metrics in the second-tier group suggest that our approach prioritizes semantic preservation and cross-task coherence.

To complement the PSNR- and SSIM-based analysis, we evaluate all task pairs whose target task is colorization, style transfer, or deraining using perceptual, distributional, and color-aware metrics. Tables~\ref{tab:new-metrics-main} and~\ref{tab:new-metrics-extra} report Fixed and Ours side by side for each backbone, where lower values indicate better agreement with the task-$B$ ground truth or target distribution. CIEDE2000 is used only for colorization and style-transfer targets, ArtFID only for style-transfer targets, and deraining-target rows are evaluated with LPIPS, DISTS, and FID.

\subsection{Single-Shot Cross-Task Results}
\label{sec:single-shot}

We additionally evaluate Gemini under the single-shot setting, performing one generation per query for both the fixed prompt and the Qwen prompt and computing all metrics on that output without result selection.
Here, the groups are defined separately for the single-shot protocol, with top-tier pairs improving all three metrics and second-tier pairs improving VIEScore but not all three metrics. The definition does not match that of the best-of-10 protocol.

\paragraph{Cross-Task Results in Top-Tier Group.}
Table~\ref{tab:2-top} presents quantitative comparisons across fourteen representative task pairs on Gemini, each compared with its fixed-prompt baseline.
Among these tasks, PSNR, SSIM, and VIEScore all consistently outperform the fixed-prompt baseline. Notably, VIEScore improves throughout, highlighting the effectiveness of our implicit prompt-driven generalization mechanism.

\paragraph{Cross-Task Results in Second-Tier Group.}
Across another ten cross-task pairs in Table~\ref{tab:2-second}, our method consistently improves VIEScore over the fixed-prompt baseline. The other two IQA metrics have slight declines in several cases. Interestingly, for these compositions, improvements are primarily reflected in the VIEScore rather than in the traditional fidelity metrics.

\begin{table*}[t]
\centering
\caption{Average of single-shot evaluation metrics. Generated by Gemini with Qwen prompts, compared to the fixed-prompt baseline. Results are divided into: (a) top-tier group, where Ours improves all three metrics; and (b) second-tier group, where VIEScore improves.}
\label{tab:2}
\begin{subtable}[t]{0.497\textwidth}
\centering
\captionsetup{skip=4pt}
\caption{Top-tier group.}
\label{tab:2-top}
\scriptsize
\setlength{\tabcolsep}{1pt}
\renewcommand{\arraystretch}{1.03}
\begin{tabularx}{\linewidth}{>{\raggedright\arraybackslash}Xcccccc}
\toprule
\multirow{2}{*}{\textbf{Task}}
& \multicolumn{2}{c}{\textbf{PSNR}$\uparrow$}
& \multicolumn{2}{c}{\textbf{SSIM}$\uparrow$}
& \multicolumn{2}{c}{\textbf{VIEScore}$\uparrow$} \\
& \textcolor{gray}{\textbf{Fixed}} & \textcolor{gray}{\textbf{Ours}}
& \textcolor{gray}{\textbf{Fixed}} & \textcolor{gray}{\textbf{Ours}}
& \textcolor{gray}{\textbf{Fixed}} & \textcolor{gray}{\textbf{Ours}} \\
\midrule
\rowcolor{lightred!30}
Deblurring $\rightarrow$ Dehazing
& 10.22 & \cellcolor{lightred!60}\textbf{11.44}
& 0.406 & \cellcolor{lightred!60}\textbf{0.461}
& 5.10 & \cellcolor{lightred!60}\textbf{5.70} \\
\grayhline
\rowcolor{lightred!30}
Deblurring $\rightarrow$ Deraining
& 16.06 & \cellcolor{lightred!60}\textbf{17.09}
& 0.401 & \cellcolor{lightred!60}\textbf{0.421}
& 4.45 & \cellcolor{lightred!60}\textbf{6.03} \\
\grayhline
\rowcolor{lightred!30}
Dehazing $\rightarrow$ Deraining
& 16.34 & \cellcolor{lightred!60}\textbf{18.16}
& 0.428 & \cellcolor{lightred!60}\textbf{0.470}
& 5.08 & \cellcolor{lightred!60}\textbf{6.20} \\
\grayhline
\rowcolor{lightred!30}
Demoireing $\rightarrow$ Dehazing
& 10.44 & \cellcolor{lightred!60}\textbf{11.57}
& 0.384 & \cellcolor{lightred!60}\textbf{0.460}
& 5.00 & \cellcolor{lightred!60}\textbf{7.23} \\
\grayhline
\rowcolor{lightred!30}
Denoising $\rightarrow$ Deblurring
& 14.20 & \cellcolor{lightred!60}\textbf{19.47}
& 0.471 & \cellcolor{lightred!60}\textbf{0.592}
& 2.10 & \cellcolor{lightred!60}\textbf{2.75} \\
\grayhline
\rowcolor{lightred!30}
Deraining $\rightarrow$ Demoireing
& 14.17 & \cellcolor{lightred!60}\textbf{15.26}
& 0.521 & \cellcolor{lightred!60}\textbf{0.566}
& 4.40 & \cellcolor{lightred!60}\textbf{7.00} \\
\grayhline
\rowcolor{lightblue!30}
Harmonization $\rightarrow$ Light Enhancement
& 8.90 & \cellcolor{lightblue!60}\textbf{9.44}
& 0.308 & \cellcolor{lightblue!60}\textbf{0.349}
& 5.13 & \cellcolor{lightblue!60}\textbf{6.23} \\
\grayhline
\rowcolor{lightblue!30}
Inpainting $\rightarrow$ Light Enhancement
& 8.20 & \cellcolor{lightblue!60}\textbf{10.33}
& 0.241 & \cellcolor{lightblue!60}\textbf{0.421}
& 4.39 & \cellcolor{lightblue!60}\textbf{8.03} \\
\grayhline
\rowcolor{lightblue!30}
Inpainting $\rightarrow$ Style Transfer
& 11.04 & \cellcolor{lightblue!60}\textbf{11.69}
& 0.306 & \cellcolor{lightblue!60}\textbf{0.373}
& 4.10 & \cellcolor{lightblue!60}\textbf{6.90} \\
\grayhline
\rowcolor{lightblue!30}
Style Transfer $\rightarrow$ Light Enhancement
& 9.79 & \cellcolor{lightblue!60}\textbf{11.59}
& 0.309 & \cellcolor{lightblue!60}\textbf{0.421}
& 5.10 & \cellcolor{lightblue!60}\textbf{6.73} \\
\grayhline
\rowcolor{lightyellow!30}
Shadow Removal $\rightarrow$ Reflection Removal
& 16.98 & \cellcolor{lightyellow!50}\textbf{19.42}
& 0.571 & \cellcolor{lightyellow!50}\textbf{0.660}
& 4.80 & \cellcolor{lightyellow!50}\textbf{5.48} \\
\grayhline
\rowcolor{lightpurple!30}
Denoising $\rightarrow$ Light Enhancement
& 7.88 & \cellcolor{lightpurple!60}\textbf{8.99}
& 0.269 & \cellcolor{lightpurple!60}\textbf{0.345}
& 4.48 & \cellcolor{lightpurple!60}\textbf{6.18} \\
\grayhline
\rowcolor{lightpurple!30}
Deraining $\rightarrow$ Style Transfer
& 11.77 & \cellcolor{lightpurple!60}\textbf{12.08}
& 0.380 & \cellcolor{lightpurple!60}\textbf{0.418}
& 5.23 & \cellcolor{lightpurple!60}\textbf{8.33} \\
\grayhline
\rowcolor{lightpurple!30}
Light Enhancement $\rightarrow$ Deraining
& 17.75 & \cellcolor{lightpurple!60}\textbf{18.77}
& 0.405 & \cellcolor{lightpurple!60}\textbf{0.509}
& 5.40 & \cellcolor{lightpurple!60}\textbf{6.38} \\
\bottomrule
\end{tabularx}
\end{subtable}
\hfill
\begin{subtable}[t]{0.497\textwidth}
\centering
\captionsetup{skip=4pt}
\caption{Second-tier group.}
\label{tab:2-second}
\scriptsize
\setlength{\tabcolsep}{1pt}
\renewcommand{\arraystretch}{1.03}
\begin{tabularx}{\linewidth}{>{\raggedright\arraybackslash}Xcccccc}
\toprule
\multirow{2}{*}{\textbf{Task}}
& \multicolumn{2}{c}{\textbf{PSNR}$\uparrow$}
& \multicolumn{2}{c}{\textbf{SSIM}$\uparrow$}
& \multicolumn{2}{c}{\textbf{VIEScore}$\uparrow$} \\
& \textcolor{gray}{\textbf{Fixed}} & \textcolor{gray}{\textbf{Ours}}
& \textcolor{gray}{\textbf{Fixed}} & \textcolor{gray}{\textbf{Ours}}
& \textcolor{gray}{\textbf{Fixed}} & \textcolor{gray}{\textbf{Ours}} \\
\midrule
\rowcolor{lightred!30}
Deblurring $\rightarrow$ Demoireing
& \cellcolor{lightred!60}\textbf{14.20} & 13.84
& \cellcolor{lightred!60}\textbf{0.572} & 0.560
& 5.23 & \cellcolor{lightred!60}\textbf{6.50} \\
\grayhline
\rowcolor{lightred!30}
Dehazing $\rightarrow$ Denoising
& \cellcolor{lightred!60}\textbf{19.58} & 18.59
& \cellcolor{lightred!60}\textbf{0.693} & 0.661
& 6.18 & \cellcolor{lightred!60}\textbf{6.98} \\
\grayhline
\rowcolor{lightred!30}
Deraining $\rightarrow$ Denoising
& \cellcolor{lightred!60}\textbf{21.14} & 16.79
& \cellcolor{lightred!60}\textbf{0.661} & 0.639
& 4.58 & \cellcolor{lightred!60}\textbf{7.45} \\
\grayhline
\rowcolor{lightblue!30}
Colorization $\rightarrow$ Harmonization
& \cellcolor{lightblue!60}\textbf{17.74} & 15.31
& \cellcolor{lightblue!60}\textbf{0.563} & 0.453
& 7.00 & \cellcolor{lightblue!60}\textbf{7.88} \\
\grayhline
\rowcolor{lightblue!30}
Colorization $\rightarrow$ Style Transfer
& 13.09 & \cellcolor{lightblue!60}\textbf{13.14}
& \cellcolor{lightblue!60}\textbf{0.444} & 0.411
& 5.10 & \cellcolor{lightblue!60}\textbf{7.63} \\
\grayhline
\rowcolor{lightblue!30}
Inpainting $\rightarrow$ Colorization
& 18.19 & \cellcolor{lightblue!60}\textbf{20.01}
& \cellcolor{lightblue!60}\textbf{0.764} & 0.657
& 4.53 & \cellcolor{lightblue!60}\textbf{6.43} \\
\grayhline
\rowcolor{lightblue!30}
Inpainting $\rightarrow$ Harmonization
& 15.85 & \cellcolor{lightblue!60}\textbf{17.02}
& \cellcolor{lightblue!60}\textbf{0.571} & 0.483
& 8.00 & \cellcolor{lightblue!60}\textbf{9.28} \\
\grayhline
\rowcolor{lightblue!30}
Light Enhancement $\rightarrow$ Colorization
& 19.57 & \cellcolor{lightblue!60}\textbf{20.08}
& \cellcolor{lightblue!60}\textbf{0.688} & 0.662
& 3.25 & \cellcolor{lightblue!60}\textbf{7.48} \\
\grayhline
\rowcolor{lightgreen!30}
Light Enhancement $\rightarrow$ Shadow Removal
& 15.33 & \cellcolor{lightgreen!60}\textbf{16.56}
& \cellcolor{lightgreen!60}\textbf{0.257} & 0.246
& 5.18 & \cellcolor{lightgreen!60}\textbf{7.15} \\
\grayhline
\rowcolor{lightorange!30}
Reflection Removal $\rightarrow$ Dehazing
& 11.03 & \cellcolor{lightorange!50}\textbf{11.11}
& \cellcolor{lightorange!50}\textbf{0.487} & 0.481
& 6.13 & \cellcolor{lightorange!50}\textbf{7.50} \\
\bottomrule
\end{tabularx}
\end{subtable}
\vspace{-6pt}
\end{table*}

\subsection{Human Preference Study}

In addition to the objective metrics and VIEScores above, we conducted a method-blind human preference study on 200 pairs generated by Gemini, comparing our method against the Fixed baseline. During the survey, prompt identities were masked, the output display position was randomized, and the ground-truth target was shown as reference.
For each item, the annotator selected the output that better matched the target transformation while preserving relevant content, with ties allowed.
Table~\ref{tab:human-preference} shows that our T2T-VICL receives a higher preference rate, and this provides a non-VLM check on the aggregate comparison.

\begin{table}[ht]
\small
\begin{minipage}[t]{\columnwidth}
\vspace{0pt}
\centering
\caption{Method-blind human preference on 200 randomly selected single-shot comparisons.}
\label{tab:human-preference}
\begin{tabular*}{\columnwidth}{@{\extracolsep{\fill}}lrr}\toprule
Choice & Count & Percentage \\
\midrule
Ours & 118 & 59.0\% \\
Fixed & 36 & 18.0\% \\
Tie & 46 & 23.0\% \\
\midrule
Ours among decisive cases & 118/154 & 76.6\% \\
\bottomrule
\end{tabular*}
\end{minipage}
\vspace{-12pt}
\end{table}

\subsection{Cross-Task Results of Other Models}
The generalization of our pipeline to other image-editing models is presented in Table~\ref{tab:5}. We include additional open-source models for evaluation, including FLUX.2-dev~\cite{black2025flux}, OmniGen 2~\cite{wu2025omnigen2}, Qwen-Image-Edit~\cite{wu2025qwen, zhao2026qwen}, and FireRed~\cite{team2026firered}. These model choices stem from the VICL setting, in which VLMs accept multiple images and a textual prompt as input to generate a target image. At present, a limited number of any-to-any VLMs support this functionality. These selected VLMs are capable of handling multi-image inputs and generating edited images as outputs. Although different models may exhibit varying strengths across specific tasks, all evaluated models demonstrate a certain degree of cross-task generalization. In terms of overall performance, Gemini-2.5-Flash achieves the highest number of top-tier and second-tier cases across different task-transfer pairs. This observation further validates our choice of Gemini as the final image generation backbone in our framework.

\begin{table*}[ht]
\centering
\scriptsize
\setlength{\tabcolsep}{0.5pt}
\renewcommand{\arraystretch}{0.88}
\captionsetup{width=0.98\textwidth, skip=4pt, type=table}
\caption{Average of evaluation metrics for additional backbones, generated by open-source models (FLUX 2, OmniGen 2, Qwen-Image, and FireRed) with fixed prompts and Qwen prompts.}
\vspace{-2pt}
\label{tab:5}

\begin{subtable}[t]{\textwidth}
\centering
\caption{FLUX.2-dev and OmniGen 2}
\vspace{-3pt}
\begin{adjustbox}{width=\textwidth}
\begin{tabular}{>{\raggedright\arraybackslash}m{4cm} *{5}{>{\centering\arraybackslash}m{0.8cm}} >{\centering\arraybackslash}m{1cm} *{5}{>{\centering\arraybackslash}m{0.8cm}} >{\centering\arraybackslash}m{1cm}}
\toprule
\multirow{3}{*}{\textbf{Task}}
& \multicolumn{6}{c}{\textbf{FLUX.2-dev}}
& \multicolumn{6}{c}{\textbf{OmniGen 2}} \\
& \multicolumn{2}{c}{\textbf{PSNR}$\uparrow$}
& \multicolumn{2}{c}{\textbf{SSIM}$\uparrow$}
& \multicolumn{2}{c}{\textbf{VIEScore}$\uparrow$}
& \multicolumn{2}{c}{\textbf{PSNR}$\uparrow$}
& \multicolumn{2}{c}{\textbf{SSIM}$\uparrow$}
& \multicolumn{2}{c}{\textbf{VIEScore}$\uparrow$} \\
& \textcolor{gray}{\textbf{Fixed}} & \textcolor{gray}{\textbf{Ours}}
& \textcolor{gray}{\textbf{Fixed}} & \textcolor{gray}{\textbf{Ours}}
& \textcolor{gray}{\textbf{Fixed}} & \textcolor{gray}{\textbf{Ours}}
& \textcolor{gray}{\textbf{Fixed}} & \textcolor{gray}{\textbf{Ours}}
& \textcolor{gray}{\textbf{Fixed}} & \textcolor{gray}{\textbf{Ours}}
& \textcolor{gray}{\textbf{Fixed}} & \textcolor{gray}{\textbf{Ours}} \\
\midrule
\rowcolor{lightred!30}
Deblurring $\rightarrow$ Dehazing
& \cellcolor{lightred!60}8.53 & 8.50
& 0.196 & \cellcolor{lightred!60}0.217
& 2.61 & \cellcolor{lightred!60}\tabgainsubscript{3.88}{+1.27}
& 7.46 & \cellcolor{lightred!60}7.62
& 0.175 & \cellcolor{lightred!60}0.183
& 2.72 & \cellcolor{lightred!60}\tabgainsubscript{3.41}{+0.69} \\
\grayhline
\rowcolor{lightred!30}
Dehazing $\rightarrow$ Denoising
& \cellcolor{lightred!60}8.81 & 8.70
& \cellcolor{lightred!60}0.498 & 0.483
& 4.68 & \cellcolor{lightred!60}\tabgainsubscript{4.77}{+0.09}
& 6.89 & \cellcolor{lightred!60}7.87
& \cellcolor{lightred!60}0.404 & 0.403
& 2.61 & \cellcolor{lightred!60}\tabgainsubscript{2.71}{+0.10} \\
\grayhline
\rowcolor{lightred!30}
Deraining $\rightarrow$ Demoireing
& \cellcolor{lightred!60}9.51 & 8.99
& \cellcolor{lightred!60}0.461 & 0.444
& \cellcolor{lightred!60}5.05 & \tablosssubscript{5.03}{-0.02}
& 7.40 & \cellcolor{lightred!60}7.55
& 0.373 & \cellcolor{lightred!60}0.375
& 2.66 & \cellcolor{lightred!60}\tabgainsubscript{2.98}{+0.32} \\
\grayhline
\rowcolor{lightblue!30}
Harmonization $\rightarrow$ Style Transfer
& \cellcolor{lightblue!60}9.07 & 8.83
& \cellcolor{lightblue!60}0.253 & 0.246
& 2.95 & \cellcolor{lightblue!60}\tabgainsubscript{5.68}{+2.73}
& \cellcolor{lightblue!60}8.44 & 7.84
& \cellcolor{lightblue!60}0.251 & 0.227
& 5.03 & \cellcolor{lightblue!60}\tabsamesubscript{5.03}{+0.00} \\
\grayhline
\rowcolor{lightblue!30}
Inpainting $\rightarrow$ Style Transfer
& \cellcolor{lightblue!60}9.02 & 8.80
& \cellcolor{lightblue!60}0.201 & 0.189
& 2.26 & \cellcolor{lightblue!60}\tabgainsubscript{3.78}{+1.52}
& \cellcolor{lightblue!60}8.30 & 7.76
& \cellcolor{lightblue!60}0.208 & 0.185
& 3.87 & \cellcolor{lightblue!60}\tabgainsubscript{4.10}{+0.23} \\
\grayhline
\rowcolor{lightyellow!30}
Shadow Removal $\rightarrow$ Reflection Removal
& 8.08 & \cellcolor{lightyellow!60}11.73
& 0.251 & \cellcolor{lightyellow!60}0.344
& 2.79 & \cellcolor{lightyellow!60}\tabgainsubscript{5.37}{+2.58}
& 8.45 & \cellcolor{lightyellow!60}8.99
& \cellcolor{lightyellow!60}0.226 & 0.215
& 2.60 & \cellcolor{lightyellow!60}\tabgainsubscript{3.16}{+0.56} \\
\grayhline
\rowcolor{lightpurple!30}
Denoising $\rightarrow$ Light Enhancement
& \cellcolor{lightpurple!60}9.63 & 8.56
& \cellcolor{lightpurple!60}0.260 & 0.193
& 7.12 & \cellcolor{lightpurple!60}\tabgainsubscript{7.62}{+0.50}
& \cellcolor{lightpurple!60}8.57 & 7.50
& \cellcolor{lightpurple!60}0.218 & 0.146
& 6.62 & \cellcolor{lightpurple!60}\tabgainsubscript{7.08}{+0.46} \\
\grayhline
\rowcolor{lightorange!30}
Reflection Removal $\rightarrow$ Dehazing
& 8.77 & \cellcolor{lightorange!50}9.45
& 0.276 & \cellcolor{lightorange!50}0.308
& 4.49 & \cellcolor{lightorange!50}\tabgainsubscript{5.88}{+1.39}
& \cellcolor{lightorange!50}7.40 & 7.30
& \cellcolor{lightorange!50}0.241 & 0.236
& 4.12 & \cellcolor{lightorange!50}\tabgainsubscript{4.88}{+0.76} \\
\grayhline
\rowcolor{lightorange!30}
Shadow Removal $\rightarrow$ Deraining
& 9.00 & \cellcolor{lightorange!50}11.80
& 0.230 & \cellcolor{lightorange!50}0.261
& 2.67 & \cellcolor{lightorange!50}\tabgainsubscript{4.44}{+1.77}
& 8.02 & \cellcolor{lightorange!50}8.48
& \cellcolor{lightorange!50}0.146 & 0.134
& 2.00 & \cellcolor{lightorange!50}\tabgainsubscript{2.71}{+0.71} \\
\bottomrule
\end{tabular}
\end{adjustbox}
\end{subtable}

\vspace{3pt}

\begin{subtable}[t]{\textwidth}
\centering
\caption{Qwen-Image-Edit-2511 and FireRed-Image-Edit-1.1}
\vspace{-3pt}
\begin{adjustbox}{width=\textwidth}
\begin{tabular}{>{\raggedright\arraybackslash}m{4cm} *{5}{>{\centering\arraybackslash}m{0.8cm}} >{\centering\arraybackslash}m{1cm} *{5}{>{\centering\arraybackslash}m{0.8cm}} >{\centering\arraybackslash}m{1cm}}
\toprule
\multirow{3}{*}{\textbf{Task}}
& \multicolumn{6}{c}{\textbf{Qwen-Image-Edit}}
& \multicolumn{6}{c}{\textbf{FireRed-Image-Edit}} \\
& \multicolumn{2}{c}{\textbf{PSNR}$\uparrow$}
& \multicolumn{2}{c}{\textbf{SSIM}$\uparrow$}
& \multicolumn{2}{c}{\textbf{VIEScore}$\uparrow$}
& \multicolumn{2}{c}{\textbf{PSNR}$\uparrow$}
& \multicolumn{2}{c}{\textbf{SSIM}$\uparrow$}
& \multicolumn{2}{c}{\textbf{VIEScore}$\uparrow$} \\
& \textcolor{gray}{\textbf{Fixed}} & \textcolor{gray}{\textbf{Ours}}
& \textcolor{gray}{\textbf{Fixed}} & \textcolor{gray}{\textbf{Ours}}
& \textcolor{gray}{\textbf{Fixed}} & \textcolor{gray}{\textbf{Ours}}
& \textcolor{gray}{\textbf{Fixed}} & \textcolor{gray}{\textbf{Ours}}
& \textcolor{gray}{\textbf{Fixed}} & \textcolor{gray}{\textbf{Ours}}
& \textcolor{gray}{\textbf{Fixed}} & \textcolor{gray}{\textbf{Ours}} \\
\midrule
\rowcolor{lightred!30}
Deblurring $\rightarrow$ Dehazing
& 8.16 & \cellcolor{lightred!60}8.56
& 0.183 & \cellcolor{lightred!60}0.238
& 3.58 & \cellcolor{lightred!60}\tabgainsubscript{5.73}{+2.15}
& \cellcolor{lightred!60}8.90 & 8.27
& \cellcolor{lightred!60}0.260 & 0.178
& 6.16 & \cellcolor{lightred!60}\tabgainsubscript{6.46}{+0.30} \\
\grayhline
\rowcolor{lightred!30}
Deblurring $\rightarrow$ Deraining
& 10.42 & \cellcolor{lightred!60}10.67
& 0.188 & \cellcolor{lightred!60}0.234
& 4.36 & \cellcolor{lightred!60}\tabgainsubscript{6.15}{+1.79}
& \cellcolor{lightred!60}9.35 & 8.92
& \cellcolor{lightred!60}0.164 & 0.139
& 5.20 & \cellcolor{lightred!60}\tabgainsubscript{6.21}{+1.01} \\
\grayhline
\rowcolor{lightred!30}
Dehazing $\rightarrow$ Denoising
& 11.91 & \cellcolor{lightred!60}15.61
& 0.487 & \cellcolor{lightred!60}0.597
& 3.71 & \cellcolor{lightred!60}\tabgainsubscript{6.10}{+2.39}
& 7.87 & \cellcolor{lightred!60}9.09
& 0.457 & \cellcolor{lightred!60}0.473
& 5.78 & \cellcolor{lightred!60}\tabgainsubscript{6.03}{+0.25} \\
\grayhline
\rowcolor{lightred!30}
Dehazing $\rightarrow$ Deraining
& 8.74 & \cellcolor{lightred!60}11.15
& 0.177 & \cellcolor{lightred!60}0.276
& 2.14 & \cellcolor{lightred!60}\tabgainsubscript{4.91}{+2.77}
& 8.56 & \cellcolor{lightred!60}8.68
& \cellcolor{lightred!60}0.193 & 0.181
& 3.06 & \cellcolor{lightred!60}\tabgainsubscript{3.94}{+0.88} \\
\grayhline
\rowcolor{lightred!30}
Demoireing $\rightarrow$ Dehazing
& 8.54 & \cellcolor{lightred!60}8.57
& 0.229 & \cellcolor{lightred!60}0.271
& 4.07 & \cellcolor{lightred!60}\tabgainsubscript{5.31}{+1.24}
& 8.71 & \cellcolor{lightred!60}9.23
& 0.276 & \cellcolor{lightred!60}0.294
& 5.41 & \cellcolor{lightred!60}\tabgainsubscript{6.16}{+0.75} \\
\grayhline
\rowcolor{lightred!30}
Denoising $\rightarrow$ Deblurring
& 9.59 & \cellcolor{lightred!60}10.36
& 0.244 & \cellcolor{lightred!60}0.295
& 4.63 & \cellcolor{lightred!60}\tabgainsubscript{7.33}{+2.70}
& 9.72 & \cellcolor{lightred!60}9.99
& 0.263 & \cellcolor{lightred!60}0.293
& 7.13 & \cellcolor{lightred!60}\tabgainsubscript{8.49}{+1.36} \\
\grayhline
\rowcolor{lightblue!30}
Colorization $\rightarrow$ Style Transfer
& \cellcolor{lightblue!60}9.24 & 9.10
& 0.245 & \cellcolor{lightblue!60}0.279
& 6.46 & \cellcolor{lightblue!60}\tabgainsubscript{6.54}{+0.08}
& \cellcolor{lightblue!60}8.75 & 8.46
& \cellcolor{lightblue!60}0.262 & 0.235
& 7.10 & \cellcolor{lightblue!60}\tabgainsubscript{7.90}{+0.80} \\
\grayhline
\rowcolor{lightblue!30}
Harmonization $\rightarrow$ Light Enhancement
& \cellcolor{lightblue!60}9.32 & 8.85
& \cellcolor{lightblue!60}0.277 & 0.230
& 7.23 & \cellcolor{lightblue!60}\tabgainsubscript{8.18}{+0.95}
& \cellcolor{lightblue!60}8.97 & 8.77
& \cellcolor{lightblue!60}0.225 & 0.205
& 8.14 & \cellcolor{lightblue!60}\tabgainsubscript{8.33}{+0.19} \\
\grayhline
\rowcolor{lightblue!30}
Inpainting $\rightarrow$ Style Transfer
& \cellcolor{lightblue!60}9.12 & 9.09
& 0.207 & \cellcolor{lightblue!60}0.231
& 4.40 & \cellcolor{lightblue!60}\tabgainsubscript{5.00}{+0.60}
& \cellcolor{lightblue!60}9.08 & 8.82
& \cellcolor{lightblue!60}0.211 & 0.198
& 4.12 & \cellcolor{lightblue!60}\tabgainsubscript{5.08}{+0.96} \\
\grayhline
\rowcolor{lightyellow!30}
Shadow Removal $\rightarrow$ Reflection Removal
& 9.09 & \cellcolor{lightyellow!60}10.78
& 0.161 & \cellcolor{lightyellow!60}0.318
& 3.06 & \cellcolor{lightyellow!60}\tabgainsubscript{3.93}{+0.87}
& \cellcolor{lightyellow!60}9.11 & 8.43
& \cellcolor{lightyellow!60}0.226 & 0.210
& 3.35 & \cellcolor{lightyellow!60}\tabgainsubscript{3.55}{+0.20} \\
\grayhline
\rowcolor{lightpurple!30}
Denoising $\rightarrow$ Light Enhancement
& \cellcolor{lightpurple!60}9.09 & 7.99
& \cellcolor{lightpurple!60}0.252 & 0.183
& 5.83 & \cellcolor{lightpurple!60}\tabgainsubscript{6.76}{+0.93}
& \cellcolor{lightpurple!60}9.60 & 9.22
& \cellcolor{lightpurple!60}0.254 & 0.224
& 8.64 & \cellcolor{lightpurple!60}\tabgainsubscript{8.70}{+0.06} \\
\grayhline
\rowcolor{lightpurple!30}
Light Enhancement $\rightarrow$ Deraining
& 9.51 & \cellcolor{lightpurple!60}10.07
& 0.203 & \cellcolor{lightpurple!60}0.231
& 3.47 & \cellcolor{lightpurple!60}\tabgainsubscript{3.48}{+0.01}
& \cellcolor{lightpurple!60}9.08 & 8.80
& \cellcolor{lightpurple!60}0.191 & 0.177
& 5.23 & \cellcolor{lightpurple!60}\tabgainsubscript{5.41}{+0.18} \\
\grayhline
\rowcolor{lightorange!30}
Reflection Removal $\rightarrow$ Dehazing
& \cellcolor{lightorange!50}8.55 & 8.46
& 0.251 & \cellcolor{lightorange!50}0.270
& 4.11 & \cellcolor{lightorange!50}\tabgainsubscript{6.19}{+2.08}
& \cellcolor{lightorange!50}8.53 & 8.18
& \cellcolor{lightorange!50}0.277 & 0.240
& 6.50 & \cellcolor{lightorange!50}\tabgainsubscript{7.55}{+1.05} \\
\grayhline
\rowcolor{lightorange!30}
Shadow Removal $\rightarrow$ Deraining
& 9.89 & \cellcolor{lightorange!50}10.44
& 0.165 & \cellcolor{lightorange!50}0.213
& 3.71 & \cellcolor{lightorange!50}\tabsamesubscript{3.71}{+0.00}
& \cellcolor{lightorange!50}10.01 & 8.87
& \cellcolor{lightorange!50}0.191 & 0.173
& 3.85 & \cellcolor{lightorange!50}\tabgainsubscript{4.07}{+0.22} \\
\bottomrule
\end{tabular}
\end{adjustbox}
\end{subtable}

\vspace{-3pt}
\end{table*}

\noindent \textbf{Computational cost comparison.}
Table~\ref{tab:efficiency} reports the computational cost of T2T-VICL with Qwen-Image-Edit, FLUX.2-dev, OmniGen 2, and FireRed-Image-Edit. We use the original task $A$ and $B$ pairs and measure ten distinct queries per task pair. Each model is loaded once and measured sequentially with batch size and concurrency one on one A100-80GB GPU. Latency uses synchronized wall-clock time and excludes model loading, while peak memory is maximum allocated PyTorch GPU memory.
For a controlled cross-backbone comparison, all editor inputs are resized to $448\times448$ resolution, different from the $224\times224$ query setting used in the main experiments.

\begin{table*}[t]
\centering
\scriptsize
\setlength{\tabcolsep}{2.6pt}
\caption{Computational cost of T2T-VICL at $448\times448$. Latency averages ten distinct queries from each of the task $A$ and $B$ pairs, and FLOPs average five queries per task pair. Loading is excluded, while prompt generation, editor inference, and postprocessing are included. Parameters include the student and image editing backbone, and Trained Parameters count only the student parameters. Peak GPU is the maximum allocated memory.}
\label{tab:efficiency}
\resizebox{\textwidth}{!}{%
\begin{tabular}{lccccccc}
\toprule
Editing Backbone & Precision & Steps & Params (B) & Trained Params (B) & FLOPs ($10^{15}$) & Latency (s) & Peak GPU (GiB) \\
\midrule
Qwen-Image-Edit & BF16 & 40 & 33.29 & 4.05 & 26.60 & 239.27 & 64.13 \\
FLUX.2-dev & BF16 & 30 & 60.76 & 4.05 & 6.61 & 65.48 & 26.51 \\
OmniGen 2 & BF16 & 50 & 12.24 & 4.05 & 2.76 & 31.50 & 23.69 \\
FireRed-Image-Edit & BF16 & 40 & 33.29 & 4.05 & 26.50 & 156.94 & 64.13 \\
\bottomrule
\end{tabular}%
}
\vspace{-6pt}
\end{table*}

\begin{table*}[t]
\begin{minipage}[t]{0.48\textwidth}
\vspace{0pt}
\centering
\captionsetup{width=0.98\linewidth, skip=4pt, type=table}
\captionof{table}{Repeated-generation analysis on Gemini. Statistics are first computed over ten generations per query and then averaged over queries. The final SSIM is paired with the PSNR-selected output.}
\label{tab:generation-variability}
\footnotesize
\renewcommand{\arraystretch}{1.0}
\begin{tabular*}{\linewidth}{@{\extracolsep{\fill}}lcccc@{}}
\toprule
& \multicolumn{2}{c}{PSNR$\uparrow$} & \multicolumn{2}{c}{SSIM$\uparrow$} \\
Protocol & Fixed & Ours & Fixed & Ours \\
\midrule
First shot & 12.788 & \textbf{13.699} & 0.417 & \textbf{0.449} \\
Mean over 10 & 12.523 & \textbf{13.553} & 0.407 & \textbf{0.446} \\
Median over 10 & 12.564 & \textbf{13.551} & 0.408 & \textbf{0.446} \\
Within-query SD$\downarrow$ & 1.372 & \textbf{1.181} & 0.068 & \textbf{0.059} \\
\arrayrulecolor{gray}
\addlinespace[1pt]
\hdashline
\addlinespace[1pt]
\arrayrulecolor{black}
PSNR-best of 10 & 14.639 & \textbf{15.384} & 0.493 & \textbf{0.527} \\
\bottomrule
\end{tabular*}
\end{minipage}
\hfill
\begin{minipage}[t]{0.5\textwidth}
\vspace{0pt}
\centering
\captionsetup{width=0.96\linewidth, skip=4pt, type=table}
\captionof{table}{Single-shot textual-guidance controls on the Gemini results. The last two rows reveal target information unavailable in our problem setting and are only diagnostic upper references.}
\label{tab:prompt-controls}
\footnotesize
\renewcommand{\arraystretch}{1.395}
\begin{tabular*}{\linewidth}{@{\extracolsep{\fill}}lccc@{}}
\toprule
Prompt & PSNR$\uparrow$ & SSIM$\uparrow$ & VIEScore$\uparrow$ \\
\midrule
Fixed structural prompt & 12.788 & 0.417 & 5.714 \\
Learned implicit prompt (ours) & 13.699 & 0.449 & 6.435 \\
\arrayrulecolor{gray}
\addlinespace[1pt]
\hdashline
\addlinespace[1pt]
\arrayrulecolor{black}
Explicit source/target names & 13.835 & 0.446 & 6.636 \\
Hand-written target description & \textbf{13.899} & \textbf{0.462} & \textbf{7.266} \\
\bottomrule
\end{tabular*}
\end{minipage}
\vspace{-6pt}
\end{table*}

\subsection{Same-Task Exploration Experiment}
We additionally report a reference experiment under the traditional same-task VICL setting.
Specifically, we conduct studies on Gemini to validate the performance gains of T2T-VICL by (1) applying VICL to the same task with a fixed prompt, and (2) applying VICL to the same task with an implicitly enriched prompt derived from same-task samples using a pipeline similar to T2T-VICL.
The enhanced prompt here is not identical to the cross-task prompt used in the main experiments, because same-task VICL has no \emph{Task Differences} component. As provided in Sec.~\ref{sec:prompt}, it instead describes only the image content and the observed visual change.
Table~\ref{tab:6} reports the results of same-task VICL with fixed and enhanced prompts.
Experimental results demonstrate improvements in PSNR, SSIM, and VIEScore across multiple task pairs.

\par\noindent\textbf{Note.} Denoising is omitted from Table~\ref{tab:6} because the SIDD images used in our same-task evaluation are large and could not be processed reliably by Gemini under this setup. The omission is therefore due to a closed-source model API processing constraint rather than a dataset construction issue.

\subsection{Additional Reliability Checks}
\label{sec:reliability}

\subsubsection{Robustness to Repeated Generation}

As closed-source models can return different outputs across repeated calls, here we test whether the advantage of the learned implicit prompt depends on a favorable single draw or on ground-truth-based candidate selection.
For the same queries from the same selected task pairs, Gemini produces ten independent outputs per query for both prompts under identical settings. We first compute the first-shot result, ten-generation mean, median, and standard deviation within each query and then average over queries. We also report the PSNR-best output and its associated SSIM as reference.

As shown in Table~\ref{tab:generation-variability}, our method is better than the fixed-prompt baseline for the first shot, the ten-generation mean, the median, and the best-of-10 result, with lower average within-query variation. Therefore, the improvement is present across repeated samples and is not explained only by selecting the highest-PSNR output.

\vspace{6pt}
\noindent
\begin{minipage}{\columnwidth}
\centering
\captionsetup{width=0.99\columnwidth, skip=4pt, type=table}
\captionof{table}{Average of metrics for same-task experiments on effects of prompt enhancement and ICL, generated by Gemini.}
\label{tab:6}
\scriptsize
\setlength{\tabcolsep}{2pt}
\renewcommand{\arraystretch}{0.95}
\begin{tabularx}{\columnwidth}{l*{6}{C}}
\toprule
\multirow{2}{*}{\textbf{Task}} 
& \multicolumn{3}{c}{\textbf{Same-task ICL + Fixed}} 
& \multicolumn{3}{c}{\textbf{Same-task ICL + Enh.}} \\
& PSNR & SSIM & VIE
& PSNR & SSIM & VIE \\
\midrule

\rowcolor{lightred!30}
Deblurring
& 17.51 & 0.529 & \cellcolor{lightred!60}4.69
& \cellcolor{lightred!60}19.29 & \cellcolor{lightred!60}0.578 & 4.30 \\

\rowcolor{lightred!30}
Dehazing
& 12.47 & 0.510 & 6.78
& \cellcolor{lightred!60}12.68 & \cellcolor{lightred!60}0.511 & \cellcolor{lightred!60}6.92 \\

\rowcolor{lightred!30}
Demoireing
& 14.98 & 0.574 & 7.41
& \cellcolor{lightred!60}15.89 & \cellcolor{lightred!60}0.596 & \cellcolor{lightred!60}7.57 \\

\rowcolor{lightred!30}
Deraining
& 19.61 & 0.563 & 7.31
& \cellcolor{lightred!60}19.86 & \cellcolor{lightred!60}0.575 & \cellcolor{lightred!60}8.00 \\

\rowcolor{lightblue!30}
Colorization
& \cellcolor{lightblue!60}21.18 & \cellcolor{lightblue!60}0.788 & 6.51
& 20.68 & 0.755 & \cellcolor{lightblue!60}7.29 \\

\rowcolor{lightblue!30}
Harmonization
& 19.88 & 0.648 & \cellcolor{lightblue!60}6.65
& \cellcolor{lightblue!60}20.18 & \cellcolor{lightblue!60}0.662 & 6.61 \\

\rowcolor{lightblue!30}
Inpainting
& 18.18 & 0.544 & 8.31
& \cellcolor{lightblue!60}18.81 & \cellcolor{lightblue!60}0.547 & \cellcolor{lightblue!60}8.65 \\

\rowcolor{lightblue!30}
Light Enhancement
& 12.27 & 0.482 & 8.26
& \cellcolor{lightblue!60}14.00 & \cellcolor{lightblue!60}0.574 & \cellcolor{lightblue!60}8.92 \\

\rowcolor{lightblue!30}
Style Transfer
& 12.96 & \cellcolor{lightblue!60}0.452 & \cellcolor{lightblue!60}6.69
& \cellcolor{lightblue!60}13.08 & 0.446 & 6.42 \\

\rowcolor{lightyellow!30}
Reflection Removal
& 20.13 & 0.718 & 5.84
& \cellcolor{lightyellow!60}22.25 & \cellcolor{lightyellow!60}0.766 & \cellcolor{lightyellow!60}5.90 \\

\rowcolor{lightyellow!30}
Shadow Removal
& \cellcolor{lightyellow!60}18.72 & \cellcolor{lightyellow!60}0.334 & 7.47
& 18.58 & 0.333 & \cellcolor{lightyellow!60}8.31 \\

\bottomrule
\end{tabularx}
\end{minipage}

\subsubsection{Effect of Textual Guidance}

This experiment explores whether the improvement comes from the learned implicit prompt rather than only from changes to the visual inputs or repeated generation.
We compare four single-shot prompt conditions on the same queries and image order, using the same evaluation metrics: the fixed structural prompt, our learned implicit prompt, explicit source and target task names, and a hand-written description of the target operation.
The last two conditions reveal target information that is usually unavailable in our setting, so they are only diagnostic upper references.
Table~\ref{tab:prompt-controls} shows that our learned prompt improves over the fixed prompt in PSNR, SSIM, and VIEScore, with its performance approaching explicit task-name guidance.
These results, to some extent, isolate the effect of learned textual guidance and show the remaining room for improving prompt prediction.

\subsubsection{Evaluation on Unseen Images}
\label{sec:unseen-images}

This analysis examines whether the performance gains from the Qwen prompt can only be attributed to reusing images seen during student model training.
Although no complete Task $A$ and $B$ input tuple in the student training data is repeated during evaluation, here we impose a stricter condition that none of the demonstration or query images appeared in the student training data before.
Across the 40 randomly selected first-shot queries satisfying this condition, spanning 14 task-pair types, our method outperforms the Fixed baseline across all metrics, achieving 14.319 versus 12.611 in PSNR, 0.462 versus 0.434 in SSIM, and 6.295 versus 5.077 in VIEScore. The Qwen prompt therefore retains its advantage on images excluded from student training.

\begin{table*}[t]
\centering
\scriptsize
\setlength{\tabcolsep}{2.8pt}
\caption{Additional evaluation metrics for targeted task pairs on Gemini ground-truth-based best-of-10 outputs. All metrics are lower-is-better. Within each Fixed/Ours pair, the lower value is shown in bold.}
\label{tab:new-metrics-main}
\resizebox{\textwidth}{!}{%
\begin{tabular}{lcccccccccc}
\toprule
\multicolumn{11}{c}{\textbf{Gemini 2.5 Flash}} \\
\midrule
\multirow{2}{*}{Task Pair} & \multicolumn{2}{c}{CIEDE$\downarrow$} & \multicolumn{2}{c}{LPIPS$\downarrow$} & \multicolumn{2}{c}{DISTS$\downarrow$} & \multicolumn{2}{c}{FID$\downarrow$} & \multicolumn{2}{c}{ArtFID$\downarrow$} \\
\cmidrule(lr){2-3}\cmidrule(lr){4-5}\cmidrule(lr){6-7}\cmidrule(lr){8-9}\cmidrule(lr){10-11}
 & Fixed & Ours & Fixed & Ours & Fixed & Ours & Fixed & Ours & Fixed & Ours \\
\midrule
\rowcolor{lightred!30}
Deblurring $\rightarrow$ Deraining & -- & -- & 0.226 & \textbf{0.194} & 0.158 & \textbf{0.138} & 79.3 & \textbf{63.3} & -- & -- \\
\rowcolor{lightred!30}
Dehazing $\rightarrow$ Deraining & -- & -- & 0.219 & \textbf{0.204} & 0.150 & \textbf{0.139} & 80.3 & \textbf{73.6} & -- & -- \\
\rowcolor{lightblue!30}
Colorization $\rightarrow$ Style Transfer & \textbf{18.56} & 19.81 & 0.522 & \textbf{0.520} & 0.242 & \textbf{0.239} & 98.5 & \textbf{97.5} & \textbf{100.2} & 102.4 \\
\rowcolor{lightblue!30}
Harmonization $\rightarrow$ Style Transfer & \textbf{19.00} & 19.50 & 0.550 & \textbf{0.530} & 0.264 & \textbf{0.246} & 117.5 & \textbf{102.4} & 105.8 & \textbf{100.1} \\
\rowcolor{lightblue!30}
Inpainting $\rightarrow$ Colorization & 10.63 & \textbf{10.60} & 0.247 & \textbf{0.209} & 0.165 & \textbf{0.144} & 97.9 & \textbf{80.8} & -- & -- \\
\rowcolor{lightblue!30}
Inpainting $\rightarrow$ Style Transfer & 28.87 & \textbf{19.23} & 0.796 & \textbf{0.508} & 0.443 & \textbf{0.236} & 274.4 & \textbf{99.3} & 284.5 & \textbf{98.3} \\
\rowcolor{lightblue!30}
Light Enhancement $\rightarrow$ Colorization & \textbf{9.83} & 10.24 & 0.219 & \textbf{0.206} & 0.152 & \textbf{0.141} & 91.8 & \textbf{78.4} & -- & -- \\
\rowcolor{lightpurple!30}
Deraining $\rightarrow$ Style Transfer & \textbf{18.44} & 19.18 & 0.574 & \textbf{0.524} & 0.269 & \textbf{0.236} & 121.2 & \textbf{99.8} & 126.0 & \textbf{103.9} \\
\rowcolor{lightpurple!30}
Light Enhancement $\rightarrow$ Deraining & -- & -- & 0.274 & \textbf{0.179} & 0.189 & \textbf{0.127} & 109.0 & \textbf{58.2} & -- & -- \\
\rowcolor{lightorange!30}
Shadow Removal $\rightarrow$ Deraining & -- & -- & 0.202 & \textbf{0.198} & 0.143 & \textbf{0.139} & 69.3 & \textbf{64.7} & -- & -- \\
\bottomrule
\end{tabular}%
}
\end{table*}

\begin{table*}[t]
\centering
\scriptsize
\setlength{\tabcolsep}{2.8pt}
\renewcommand{\arraystretch}{0.95}
\caption{Additional evaluation metrics for targeted task pairs. Open-source models use single-shot outputs while Seedream uses ground-truth-based best-of-10 outputs. All metrics are lower-is-better. Within each Fixed/Ours pair, the lower value is shown in bold.}
\label{tab:new-metrics-extra}
\resizebox{\textwidth}{!}{%
\begin{tabular}{lcccccccccc}
\toprule
\multicolumn{11}{c}{\textbf{FLUX.2-dev}} \\
\midrule
\multirow{2}{*}{Task Pair} & \multicolumn{2}{c}{CIEDE$\downarrow$} & \multicolumn{2}{c}{LPIPS$\downarrow$} & \multicolumn{2}{c}{DISTS$\downarrow$} & \multicolumn{2}{c}{FID$\downarrow$} & \multicolumn{2}{c}{ArtFID$\downarrow$} \\
\cmidrule(lr){2-3}\cmidrule(lr){4-5}\cmidrule(lr){6-7}\cmidrule(lr){8-9}\cmidrule(lr){10-11}
 & Fixed & Ours & Fixed & Ours & Fixed & Ours & Fixed & Ours & Fixed & Ours \\
\midrule
\rowcolor{lightred!30}
Deblurring $\rightarrow$ Deraining & -- & -- & 0.765 & \textbf{0.750} & 0.373 & \textbf{0.359} & 244.7 & \textbf{222.2} & -- & -- \\
\rowcolor{lightred!30}
Dehazing $\rightarrow$ Deraining & -- & -- & \textbf{0.721} & 0.756 & 0.346 & \textbf{0.340} & 236.0 & \textbf{234.9} & -- & -- \\
\rowcolor{lightblue!30}
Harmonization $\rightarrow$ Style Transfer & \textbf{30.95} & 31.91 & 0.823 & \textbf{0.812} & 0.426 & \textbf{0.421} & 252.3 & \textbf{241.0} & 267.6 & \textbf{246.3} \\
\rowcolor{lightblue!30}
Inpainting $\rightarrow$ Colorization & 33.94 & \textbf{32.62} & 0.806 & \textbf{0.778} & 0.424 & \textbf{0.390} & 242.8 & \textbf{233.7} & -- & -- \\
\rowcolor{lightblue!30}
Inpainting $\rightarrow$ Style Transfer & \textbf{31.54} & 32.25 & \textbf{0.807} & 0.807 & 0.437 & \textbf{0.434} & 245.5 & \textbf{243.4} & 276.9 & \textbf{267.2} \\
\rowcolor{lightblue!30}
Light Enhancement $\rightarrow$ Colorization & \textbf{29.89} & 33.02 & 0.773 & \textbf{0.764} & 0.382 & \textbf{0.354} & 260.3 & \textbf{243.3} & -- & -- \\
\rowcolor{lightorange!30}
Shadow Removal $\rightarrow$ Deraining & -- & -- & 0.702 & \textbf{0.630} & 0.369 & \textbf{0.325} & 234.6 & \textbf{190.7} & -- & -- \\
\end{tabular}%
}

\resizebox{\textwidth}{!}{%
\begin{tabular}{lcccccccccc}
\toprule
\multicolumn{11}{c}{\textbf{OmniGen 2}} \\
\midrule
\multirow{2}{*}{Task Pair} & \multicolumn{2}{c}{CIEDE$\downarrow$} & \multicolumn{2}{c}{LPIPS$\downarrow$} & \multicolumn{2}{c}{DISTS$\downarrow$} & \multicolumn{2}{c}{FID$\downarrow$} & \multicolumn{2}{c}{ArtFID$\downarrow$} \\
\cmidrule(lr){2-3}\cmidrule(lr){4-5}\cmidrule(lr){6-7}\cmidrule(lr){8-9}\cmidrule(lr){10-11}
 & Fixed & Ours & Fixed & Ours & Fixed & Ours & Fixed & Ours & Fixed & Ours \\
\midrule
\rowcolor{lightred!30}
Deblurring $\rightarrow$ Deraining & -- & -- & \textbf{0.782} & 0.782 & 0.402 & \textbf{0.392} & 266.4 & \textbf{249.7} & -- & -- \\
\rowcolor{lightred!30}
Dehazing $\rightarrow$ Deraining & -- & -- & 0.807 & \textbf{0.802} & 0.404 & \textbf{0.372} & \textbf{299.9} & 306.1 & -- & -- \\
\rowcolor{lightblue!30}
Harmonization $\rightarrow$ Style Transfer & \textbf{32.30} & 34.19 & 0.822 & \textbf{0.819} & 0.453 & \textbf{0.438} & 265.0 & \textbf{258.8} & 291.2 & \textbf{268.5} \\
\rowcolor{lightblue!30}
Inpainting $\rightarrow$ Style Transfer & \textbf{32.71} & 34.00 & 0.808 & \textbf{0.796} & 0.453 & \textbf{0.429} & \textbf{266.8} & 272.8 & 281.3 & \textbf{247.4} \\
\rowcolor{lightpurple!30}
Deraining $\rightarrow$ Style Transfer & \textbf{32.91} & 34.83 & \textbf{0.808} & 0.810 & 0.426 & \textbf{0.397} & 244.1 & \textbf{214.3} & 268.3 & \textbf{237.9} \\
\rowcolor{lightorange!30}
Shadow Removal $\rightarrow$ Deraining & -- & -- & \textbf{0.764} & 0.766 & 0.418 & \textbf{0.383} & 285.3 & \textbf{270.3} & -- & -- \\
\end{tabular}%
}

\resizebox{\textwidth}{!}{%
\begin{tabular}{lcccccccccc}
\toprule
\multicolumn{11}{c}{\textbf{Qwen-Image-Edit}} \\
\midrule
\multirow{2}{*}{Task Pair} & \multicolumn{2}{c}{CIEDE$\downarrow$} & \multicolumn{2}{c}{LPIPS$\downarrow$} & \multicolumn{2}{c}{DISTS$\downarrow$} & \multicolumn{2}{c}{FID$\downarrow$} & \multicolumn{2}{c}{ArtFID$\downarrow$} \\
\cmidrule(lr){2-3}\cmidrule(lr){4-5}\cmidrule(lr){6-7}\cmidrule(lr){8-9}\cmidrule(lr){10-11}
 & Fixed & Ours & Fixed & Ours & Fixed & Ours & Fixed & Ours & Fixed & Ours \\
\midrule
\rowcolor{lightred!30}
Deblurring $\rightarrow$ Deraining & -- & -- & 0.689 & \textbf{0.653} & 0.342 & \textbf{0.317} & 178.7 & \textbf{171.2} & -- & -- \\
\rowcolor{lightred!30}
Dehazing $\rightarrow$ Deraining & -- & -- & 0.804 & \textbf{0.642} & 0.390 & \textbf{0.310} & 253.0 & \textbf{210.0} & -- & -- \\
\rowcolor{lightblue!30}
Colorization $\rightarrow$ Style Transfer & 32.33 & \textbf{31.34} & 0.815 & \textbf{0.790} & 0.416 & \textbf{0.398} & 241.2 & \textbf{239.2} & 245.0 & \textbf{228.2} \\
\rowcolor{lightblue!30}
Harmonization $\rightarrow$ Style Transfer & 34.09 & \textbf{32.00} & 0.836 & \textbf{0.801} & 0.434 & \textbf{0.414} & 250.0 & \textbf{230.7} & 261.8 & \textbf{236.9} \\
\rowcolor{lightblue!30}
Inpainting $\rightarrow$ Style Transfer & 31.86 & \textbf{30.74} & 0.815 & \textbf{0.769} & 0.426 & \textbf{0.402} & 233.2 & \textbf{214.2} & 267.2 & \textbf{229.1} \\
\rowcolor{lightpurple!30}
Deraining $\rightarrow$ Style Transfer & 29.15 & \textbf{28.97} & 0.785 & \textbf{0.764} & 0.385 & \textbf{0.375} & 191.7 & \textbf{190.1} & 213.1 & \textbf{201.8} \\
\rowcolor{lightpurple!30}
Light Enhancement $\rightarrow$ Deraining & -- & -- & 0.757 & \textbf{0.715} & 0.387 & \textbf{0.364} & 235.4 & \textbf{219.7} & -- & -- \\
\rowcolor{lightorange!30}
Shadow Removal $\rightarrow$ Deraining & -- & -- & 0.763 & \textbf{0.711} & 0.386 & \textbf{0.351} & 225.6 & \textbf{213.2} & -- & -- \\
\end{tabular}%
}

\resizebox{\textwidth}{!}{%
\begin{tabular}{lcccccccccc}
\toprule
\multicolumn{11}{c}{\textbf{FireRed-Image-Edit}} \\
\midrule
\multirow{2}{*}{Task Pair} & \multicolumn{2}{c}{CIEDE$\downarrow$} & \multicolumn{2}{c}{LPIPS$\downarrow$} & \multicolumn{2}{c}{DISTS$\downarrow$} & \multicolumn{2}{c}{FID$\downarrow$} & \multicolumn{2}{c}{ArtFID$\downarrow$} \\
\cmidrule(lr){2-3}\cmidrule(lr){4-5}\cmidrule(lr){6-7}\cmidrule(lr){8-9}\cmidrule(lr){10-11}
 & Fixed & Ours & Fixed & Ours & Fixed & Ours & Fixed & Ours & Fixed & Ours \\
\midrule
\rowcolor{lightred!30}
Dehazing $\rightarrow$ Deraining & -- & -- & 0.801 & \textbf{0.785} & 0.387 & \textbf{0.365} & 289.9 & \textbf{271.4} & -- & -- \\
\rowcolor{lightblue!30}
Harmonization $\rightarrow$ Style Transfer & \textbf{31.75} & 32.27 & 0.825 & \textbf{0.822} & 0.443 & \textbf{0.439} & 251.0 & \textbf{244.2} & 262.7 & \textbf{258.9} \\
\rowcolor{lightblue!30}
Inpainting $\rightarrow$ Style Transfer & \textbf{31.08} & 31.80 & 0.801 & \textbf{0.799} & 0.435 & \textbf{0.433} & 243.8 & \textbf{232.6} & 262.3 & \textbf{260.3} \\
\rowcolor{lightpurple!30}
Deraining $\rightarrow$ Style Transfer & \textbf{29.18} & 30.12 & 0.792 & \textbf{0.789} & \textbf{0.393} & 0.394 & 209.1 & \textbf{204.5} & 230.4 & \textbf{227.4} \\
\rowcolor{lightpurple!30}
Light Enhancement $\rightarrow$ Deraining & -- & -- & 0.801 & \textbf{0.781} & 0.403 & \textbf{0.392} & 268.0 & \textbf{250.4} & -- & -- \\
\end{tabular}%
}

\resizebox{\textwidth}{!}{%
\begin{tabular}{lcccccccccc}
\toprule
\multicolumn{11}{c}{\textbf{Seedream 4.0}} \\
\midrule
\multirow{2}{*}{Task Pair} & \multicolumn{2}{c}{CIEDE$\downarrow$} & \multicolumn{2}{c}{LPIPS$\downarrow$} & \multicolumn{2}{c}{DISTS$\downarrow$} & \multicolumn{2}{c}{FID$\downarrow$} & \multicolumn{2}{c}{ArtFID$\downarrow$} \\
\cmidrule(lr){2-3}\cmidrule(lr){4-5}\cmidrule(lr){6-7}\cmidrule(lr){8-9}\cmidrule(lr){10-11}
 & Fixed & Ours & Fixed & Ours & Fixed & Ours & Fixed & Ours & Fixed & Ours \\
\midrule
\rowcolor{lightred!30}
Deblurring $\rightarrow$ Deraining & -- & -- & 0.414 & \textbf{0.398} & 0.250 & \textbf{0.239} & 135.9 & \textbf{113.8} & -- & -- \\
\rowcolor{lightred!30}
Dehazing $\rightarrow$ Deraining & -- & -- & 0.371 & \textbf{0.357} & \textbf{0.228} & 0.229 & 119.2 & \textbf{112.5} & -- & -- \\
\rowcolor{lightblue!30}
Colorization $\rightarrow$ Style Transfer & \textbf{28.40} & 29.21 & 0.629 & \textbf{0.623} & 0.297 & \textbf{0.295} & 113.2 & \textbf{103.0} & 139.1 & \textbf{132.5} \\
\rowcolor{lightblue!30}
Light Enhancement $\rightarrow$ Colorization & 23.39 & \textbf{22.82} & 0.412 & \textbf{0.383} & 0.223 & \textbf{0.208} & 130.3 & \textbf{103.7} & -- & -- \\
\rowcolor{lightpurple!30}
Light Enhancement $\rightarrow$ Deraining & -- & -- & 0.398 & \textbf{0.364} & 0.255 & \textbf{0.228} & 131.6 & \textbf{104.4} & -- & -- \\
\bottomrule
\end{tabular}%
}
\end{table*}

\end{document}